\documentclass[11pt]{article}

\usepackage[final]{acl}

\usepackage{times}
\usepackage{latexsym}

\usepackage[T1]{fontenc}
\usepackage[utf8]{inputenc}

\usepackage{microtype}
\usepackage{inconsolata}

\usepackage{graphicx}

\usepackage{booktabs}
\usepackage{amsmath}
\usepackage{enumitem}
\usepackage{amssymb}
\usepackage{array}
\usepackage{tcolorbox}
\usepackage{multirow}
\usepackage{xcolor}
\usepackage{pifont}
\usepackage{enumitem}
\usepackage[table]{xcolor}
\usepackage{tabularx}
\usepackage{float}
\newcommand{\ours}{\texttt{SAFE}}

\title{SAFE: An LLM-as-Verifier Framework 

for Evidence-Grounded Multi-Hop Reasoning}

\author{
  Daeyong Kwon \\
  \texttt{daeyongkwon@snu.ac.kr}\\\\
  \And
  Soyoung Yoon \\
  \texttt{soyoung.yoon@snu.ac.kr}\\\\
  Seoul National University, South Korea
  \And
  Seung-won Hwang\thanks{Corresponding author}\\
  \texttt{seungwonh@snu.ac.kr}\\\\
}

\begin{document}
\maketitle

\begin{abstract}
Multi-hop QA benchmarks often reward Large Language Models (LLMs) for spurious correctness, where models reach correct answers through invalid intermediate reasoning. 
We propose \ours{}, an LLM-as-verifier framework for evidence-grounded multi-hop QA.
Rather than judging only the final answer after generation, \ours{} verifies reasoning during generation by checking intermediate steps against the provided passages and previous reasoning trajectory.
To make this process checkable, \ours{} decomposes reasoning into atomic, evidence-grounded units represented with Knowledge Graph (KG) triples.
At train-time, \ours{} verifies benchmark supervision under KG-grounded constraints and constructs reliable verifier training data. 
At inference-time, an external verifier checks each generated step, identifies invalid reasoning, and provides correction feedback before errors propagate. 
Across three multi-hop QA benchmarks, \ours{} improves accuracy by 8.8 pp on average. 
These results show that evidence-grounded multi-hop QA benefits from shifting LLM-based evaluation from post-hoc answer judgment to stepwise reasoning verification.\footnote{Code and data are available at \url{https://github.com/DaeyongKwon98/SAFE}}
\end{abstract}

\section{Introduction}\label{sec:introduction}
The frontier of Large Language Models (LLMs) has increasingly shifted toward complex reasoning tasks, such as multi-hop question answering (QA), where a model must synthesize evidence across multiple documents to derive a final answer. 
Chain-of-Thought (CoT) prompting has substantially improved performance on such tasks~\citep{wei2022chain}, but its free-form reasoning traces remain difficult to ground or verify~\citep{turpin2023language}. 
A model may arrive at the correct answer through intermediate steps that introduce unverified entities, unsupported relations, or implicit shortcuts. 
As a result, final-answer accuracy alone can reward spurious reasoning rather than evidence-grounded inference.

Recent LLM-as-judge approaches make it possible to evaluate generated answers, but judging completed outputs is not sufficient for multi-hop reasoning. 
In evidence-grounded QA, errors often arise before the final answer. 
A single unsupported entity or wrong relation can mislead the subsequent reasoning trajectory. 
This suggests that verification should move from output-level judgment to process-level verification. 
Rather than asking only whether the final answer is correct, an LLM-as-verifier framework should check whether each intermediate reasoning step is supported by the provided evidence and provide feedback before errors propagate.


We propose SAFE, an LLM-as-verifier framework for evidence-grounded multi-hop QA, characterized by three design principles. First, SAFE performs online verification: each reasoning step is checked immediately after generation and before it is committed to the subsequent reasoning trajectory. Second, SAFE uses atomic verification units: Knowledge Graph (KG) triples serve as compact, passage-verifiable representations of individual reasoning steps, rather than primarily as structures for retrieval or reasoning-path construction. Third, SAFE performs externally grounded verification: a separate verifier checks each step against the provided passages and returns structured diagnosis and correction guidance when the step is unsupported. Together, these design choices allow SAFE to detect and correct invalid intermediate reasoning before errors propagate across hops.

\ours{} operates at both train-time and inference-time. 
At train-time, \ours{} verifies benchmark supervision by enforcing atomicity and evidence grounding over standard multi-hop QA datasets~\citep{ho2020constructing,yang2018hotpotqa,trivedi2022musique}. 
This process removes up to 14\% of instances whose reasoning supervision is not fully verifiable. 
More importantly, train-time verification constructs reliable supervision for the verifier itself. 
Instead of treating noisy benchmark reasoning paths as gold, \ours{} converts unverifiable reasoning steps into targeted diagnostic and corrective signals. 
This allows the verifier to learn not only whether a step is valid, but also how to respond when the reasoning process deviates from the evidence.

At inference-time, \ours{} uses an external verifier as a stepwise process controller. 
Given a partial reasoning trajectory, the verifier checks whether the next generated step is grounded in the evidence and corresponds to a single evidence-supported KG triple. 
If the step is invalid, \ours{} localizes the error and provides correction feedback for revising the next step. 
This differs from LLM self-correction, which relies on the same model to critique its own reasoning and is often unreliable~\citep{huang2023large,tyen2024llms}. 
By separating generation from verification, \ours{} maintains evidence-grounded reasoning trajectories rather than merely scoring completed outputs.

Empirically, \ours{} significantly improves multi-hop QA performance, achieving an average +8.8 pp accuracy gain over baseline reasoning methods. 
Further analyses show that these gains come from structured step-level verification and feedback, rather than generic feedback alone. 
We also find that \ours{} remains effective across different verifier-training and generator settings, and can be extended to incomplete-evidence scenarios.

Our main contributions are as follows:
\begin{itemize}[leftmargin=*, noitemsep, topsep=0pt]
    \item We formulate evidence-grounded multi-hop QA as an LLM-as-verifier problem, shifting from post-hoc LLM-as-judge evaluation to stepwise verification of intermediate reasoning.


    \item We propose SAFE, an LLM-as-verifier framework characterized by online verification, atomic KG-style reasoning units, and external evidence grounding, enabling reliable verifier supervision at train-time and correction of invalid reasoning steps at inference-time.

    \item We show that \ours{} achieves a +8.8 pp accuracy gain, with analyses confirming the importance and robustness of structured step-level verification and feedback.
\end{itemize}

\section{Related Works}\label{sec:related_works}
\subsection{From LLM-as-Judge to LLM-as-Verifier}

Recent LLM-as-judge methods use LLMs as scalable evaluators for generated answers, rationales, and open-ended outputs~\citep{zheng2023judging, liu2023g}. 
These methods are useful for assessing completed generations, but they typically operate after the full output has already been produced. 
In evidence-grounded multi-hop QA, however, important failures can occur before the final answer is generated~\cite{ishii2024analysis}. 
An intermediate step may introduce an unsupported entity, an incorrect relation, or a shortcut that affects the subsequent reasoning process. 
In contrast, LLM-as-verifier methods aim to check whether each intermediate step is supported by the provided evidence and provide feedback before errors propagate.

Process verification methods also emphasize intermediate reasoning steps~\citep{lightman2024let}. 
However, many of these methods are developed for domains such as mathematical reasoning~\citep{wu2025enhancing}, where step correctness is not necessarily tied to retrieved textual evidence. 
In contrast, evidence-grounded multi-hop QA requires each reasoning step to be independently grounded in the given context. 
\ours{} targets this setting by treating multi-hop QA as stepwise verification over evidence-grounded reasoning steps.

\subsection{Multi-hop QA Benchmarks and Spurious Reasoning}

Multi-hop QA benchmarks such as 2WikiMultihopQA~\citep{ho2020constructing}, HotpotQA~\citep{yang2018hotpotqa}, and MuSiQue~\citep{trivedi2022musique} have been widely adopted for evaluating complex reasoning and retrieval-augmented generation (RAG) systems~\citep{jimenez2024hipporag, trivedi2023interleaving}. 
However, final-answer accuracy can overestimate true reasoning ability. 
\citet{ishii2024analysis} show that a non-trivial portion of correct answers in standard multi-hop QA datasets arise from spurious reasoning, where models reach the right answer without following a fully supported reasoning path.

To reduce or analyze such failures, prior work has incorporated KGs before or after generation. 
For example, \citet{fang2024trace} extract KG chains before generation, while \citet{nguyen2024direct} evaluate generated CoT reasoning by mapping the reasoning steps to a KG. 
\citet{wang2025dtkg} further integrates KG-based verification and refinement into multi-hop QA through dual reasoning tracks.
More closely related to our setting is REVEAL~\citep{jacovi2024chain}, which assigns correctness labels to individual reasoning steps using a specialized error taxonomy. 
Hop, Skip, and Overthink~\citep{yadav2025hop} similarly studies post-hoc failure modes to localize reasoning breakdowns.

These approaches are important for exposing unsupported or shortcut reasoning, but they mainly analyze completed reasoning trajectories or rely on static KG chains. 
\ours{} instead uses KG-grounded atomic steps as the basis for stepwise verification. 
At train-time, this supports the construction of verifiable supervision. At inference-time, it enables the verifier to check intermediate steps before the next step generation.

\subsection{Self-Correction and Inference-Time Feedback}

LLMs can be prompted to critique and revise their own outputs~\citep{madaan2023self} or accumulate reflective feedback across trials~\citep{shinn2023reflexion}. 
Methods such as Self-Refine~\citep{madaan2023self} show that self-critique can improve output quality without additional training or external models.
RISE~\citep{he2025rise} applies iterative self-exploration and self-critique to multi-hop question answering.

However, self-correction alone is unreliable for complex multi-hop reasoning. 
The same model that produced an invalid step may fail to identify the error, especially when the error depends on evidence grounding. 
Prior work shows that LLMs often struggle to correct their own reasoning flaws without external ground-truth signals~\citep{huang2023large}. 
\citet{tyen2024llms} further identify mistake finding as a key bottleneck, showing that unguided models frequently fail to locate their own logical missteps.

\ours{} differs from pure self-correction by separating generation from verification. 
Instead of asking the generator to critique its own output, \ours{} uses an external verifier to check each reasoning step against the provided evidence. 
When a step is invalid, the verifier provides feedback that helps the generator return to evidence-supported reasoning.

\section{\ours{}: An LLM-as-Verifier Framework}\label{sec:method}
\ours{} is an LLM-as-verifier framework for evidence-grounded multi-hop QA.
This section first defines the stepwise verification protocol used by \ours{}, then describes how the framework constructs verifier supervision at train-time and applies external verification during inference.

\subsection{Evidence-Grounded Stepwise Verification}
\label{sec:method_requirements}

A verifier for evidence-grounded multi-hop QA not only assigns a correctness score, but also checks whether each intermediate step is supported before the reasoning process continues. 
\ours{} addresses this by treating intermediate reasoning steps as atomic units that can be independently checked against the provided evidence.

Concretely, we represent each reasoning step as a KG triple, so that the step corresponds to a single checkable operation. 
A step is considered valid only when its entities and relation are explicitly supported by the provided passages. 
This representation allows the verifier to determine whether the current reasoning state remains grounded before the generator proceeds to the next step.

When a proposed step is not supported, the verifier returns structured feedback for correction. 
The feedback identifies the source of the failure and provides guidance that helps the generator revise the invalid step. 
In this way, \ours{} uses verification not only to judge whether a step is valid, but also to support correction during generation.

This stepwise verification process is used at both train-time and inference-time. 
At train-time, the atomic and evidence-grounded representation guides benchmark verification and verifier-data construction. 
At inference-time, the verifier applies the same evidence-grounding criterion to each proposed step and provides feedback when correction is needed.

\subsection{Atomic Reasoning Units and Verification Feedback}
\label{sec:method_taxonomy}

We define a valid reasoning trajectory as a sequence of evidence-grounded atomic steps. 
Let $\mathcal{P}$ denote the passages provided as context. 
Each reasoning step $s_i$ must correspond to a single KG triple $(e_{head}, r, e_{tail})$, where both entities and the relation are supported by $\mathcal{P}$. 
This representation decomposes multi-hop reasoning into independently checkable units, making it possible to verify each step before proceeding to the next one.
When a step is invalid, the verifier produces structured feedback that indicates why the step fails and how the generator should revise it.
We organize this feedback using four error categories: Procedural, Attribution, Logical, and Final Answer (Appendix Table~\ref{tab:error_taxonomy}).

\ours{} applies these categories in a fixed verification order. 
Procedural errors capture invalid step structure, such as loops and disconnected reasoning paths. 
After procedural validity is checked, Attribution errors identify entities or relations that are not grounded in the provided passages. 
Logical errors capture relation-level inconsistencies, where the step uses an incorrect relation even if the entities are mentioned in the context. 
Finally, Final Answer errors arise when the reasoning trajectory is locally valid but does not reach the correct answer.
The predicted category is then used to produce feedback for revising the invalid step.

\subsection{Train-Time Verification for Reliable Verifier Supervision}
\label{sec:method_train_time}

A verifier cannot be trained or evaluated reliably if the underlying benchmark supervision is not itself verifiable.
Standard multi-hop QA datasets may contain missing relations, ambiguous entity mappings, or incomplete supporting facts that allow the correct answer without a fully grounded reasoning path. 
Such cases make it difficult to determine whether an intermediate step is valid under an evidence-grounded verification.

To address this issue, \ours{} first verifies benchmark supervision under the KG-grounded step constraints defined in Section~\ref{sec:method_taxonomy}. 
This train-time verification produces a filtered benchmark set in which verifier supervision can be constructed and evaluated from evidence-supported reasoning paths.
We use gpt-oss-120b~\citep{agarwal2025gpt} as the base LLM for this pipeline, with prompts detailed in Appendix~\ref{sec:appendix_prompts}. 
The pipeline proceeds as follows.

\begin{enumerate}[leftmargin=*, itemsep=0pt, topsep=0pt]
    \item \textbf{Triple Extraction:} We prompt the LLM to extract atomic triples $(e_{head}, r, e_{tail})$ from the gold passages, treating the context as a localized KG.
    
    \item \textbf{Iterative Gleaning:} We perform up to two additional extraction rounds to recover valid triples that may have been missed in the initial stage.
    
    \item \textbf{Entity Resolution:} We merge synonymous or coreferential entities, such as ``USA'' and ``United States'', to build a normalized local KG.
    
    \item \textbf{Reasoning Path Discovery:} We search for a continuous evidence-grounded reasoning path that connects entities to the gold answer.
    
    \item \textbf{Validation:} If no valid reasoning path can be found under this protocol, the data is marked as unverifiable and removed from the verified set.
\end{enumerate}

\begin{table}[t]
\centering
\small
\resizebox{\columnwidth}{!}{
\begin{tabular}{llccc}
\toprule
\textbf{Dataset} & \textbf{Split} & \textbf{Total} & \textbf{Unverifiable} & \textbf{Ratio} \\
\midrule
\multirow{2}{*}{2Wiki} & Train & 167,454 & 7,452 & 4.45\% \\
 & Val & 12,576 & 901 & 7.16\% \\ 
\midrule
\multirow{2}{*}{HotpotQA} & Train & 90,447 & 2,376 & 2.63\% \\
 & Val & 7,345 & 227 & 3.09\% \\ 
\midrule
\multirow{2}{*}{MuSiQue} & Train & 19,902 & 2,149 & 10.80\% \\
 & Val & 2,412 & 342 & 14.18\% \\ 
\bottomrule
\end{tabular}%
}
\caption{
Statistics of instances flagged as unverifiable by our KG-grounded benchmark verification pipeline.
}
\vspace{-1em}
\label{tab:benchmark_error_statistics}
\end{table}

Appendix Figure~\ref{fig:kg_filtering_pipeline} provides an overview of the verification pipeline, and Table~\ref{tab:benchmark_error_statistics} summarizes the benchmark-level results. 
Across the examined datasets, 3\%--14\% of instances are unverifiable under our atomic grounding constraints. 
After filtering these instances, \ours{} uses the remaining examples to train and evaluate the verifier on evidence-supported reasoning paths.

\paragraph{Constructing Verifier Training Data.}

Training the verifier requires both valid reasoning trajectories and diverse invalid steps. 
Using the verified benchmark set, we construct 3.9k ideal trajectories whose steps are grounded in the provided passages. 
We then synthesize 7.6k negative examples through controlled error injection. 
Starting from valid trajectories, a teacher model introduces step-level violations aligned with the error categories. 
This produces corrupted trajectories with an error step, category, diagnosis, and correction guidance.

Synthetic perturbations provide controlled coverage of error types, but they may not fully capture the failures made by the target generator during inference. 
To reduce this gap, we use an iterative refinement strategy. 
We first train an initial verifier on the synthetic data and analyze recurring failures, such as confusing lexically similar entities or mishandling numerical comparisons. 
For each recurring failure pattern, we use the teacher model to generate additional feedback labels. 
Through this \textit{train} $\rightarrow$ \textit{analyze} $\rightarrow$ \textit{augment} cycle, we add 848 hard cases to the training pool. 
The final verifier training set contains 11.5k instances.

The overall training set statistics are shown in Appendix Figure~\ref{fig:training_dataset_statistics}. 
Additional details, including the positional distribution of error steps and dataset statistics, are provided in Appendix~\ref{sec:appendix_training_dataset}.

\subsection{Inference-Time Stepwise Verification}
\label{sec:method_inference_time}

At inference-time, \ours{} uses an external verifier to check generated reasoning steps during generation. 
Given a question $q$, passages $\mathcal{P}$, and a partial reasoning trajectory, the generator proposes the next atomic step. 
The verifier then checks whether the proposed step is grounded in the evidence and consistent with a single evidence-supported KG triple.

{
\setlength{\abovedisplayskip}{3pt}
\setlength{\belowdisplayskip}{3pt}
\setlength{\abovedisplayshortskip}{2pt}
\setlength{\belowdisplayshortskip}{2pt}

Formally, let a reasoning trajectory be represented as a sequence of atomic steps:
\[
\tau = (s_1, s_2, \dots, s_t).
\]
Given the current reasoning prefix $s_{1:t}$, the verifier $F$ predicts:
\[
(y_t, d_t, g_t) = F(q, \mathcal{P}, s_{1:t}),
\]
}
where:
\begin{itemize}[leftmargin=*, noitemsep, topsep=0pt]
    \item $y_t$ denotes whether the current step is valid or belongs to one of the error categories defined in Section~\ref{sec:method_taxonomy}.
    
    \item $d_t$ provides a natural language diagnosis of why the step is invalid or inconsistent with the evidence.
    
    \item $g_t$ provides correction guidance for revising the invalid step or generating a valid next step.
\end{itemize}

If the verifier marks the step as valid, the step is accepted and generation continues. 
If the verifier detects an error, the generator receives the feedback, revises the step, and continues from the corrected reasoning prefix. 
In this way, \ours{} verifies and corrects intermediate reasoning steps during generation.
Our inference-time diagram is described in Appendix Figure~\ref{fig:diagram}.

\section{Experiments}\label{sec:experiments}
\subsection{Experimental Setup}
\label{sec:experimental_setup}

\paragraph{Datasets.}
We evaluate \ours{} on three multi-hop QA benchmarks: 2WikiMultihopQA~\citep{ho2020constructing}, HotpotQA~\citep{yang2018hotpotqa}, and MuSiQue~\citep{trivedi2022musique}. 
We apply the KG-grounded verification pipeline in Section~\ref{sec:method_train_time} to the validation splits and construct a verified evaluation pool. 
From this pool, we sample 1,000 instances per benchmark, yielding 3,000 evaluation examples in total. 
Each question is evaluated with 10 passages, including the gold evidence passages and distractors. 
Additional data processing details are provided in Appendix~\ref{sec:appendix_evaluation_datasets}.

\paragraph{Evaluation Metrics.}
We report Exact Match (EM), token-level F1, and answer accuracy. 
EM and F1 measure surface-level overlap with the ground-truth answer, but they may underestimate performance when a generated answer is semantically correct with different wording. 
We therefore use an LLM-as-judge evaluator~\citep{li2025generation}, implemented with gpt-oss-120b, to assess answer-level semantic equivalence. 

\paragraph{Models.}
We evaluate \ours{} across multiple generator families and sizes: Qwen3-4B, Qwen3-8B~\citep{yang2025qwen3}, Qwen2.5-14B~\citep{qwen2025qwen25technicalreport}, Llama-3.1-8B~\citep{grattafiori2024llama}, and Gemma-3-12B~\citep{gemmateam2025gemma3technicalreport}.\footnote{We use the instruction-tuned version when available.} 
These models are used for the main evaluation on the full verified benchmark set. 
To further test whether \ours{} remains useful with stronger reasoning models, we additionally conduct a subset evaluation with Gemini 3.1 Flash Lite~\cite{googledeepmind2026gemini31flashlite} and GPT 5.4 mini~\cite{openai2026gpt54mini}, reported separately in Section~\ref{sec:analysis}.
The stepwise verifier is trained on the non-thinking version of Qwen3-8B to prioritize training efficiency and architectural simplicity. 
For constructing verified training data, we use gpt-oss-120b~\citep{agarwal2025gpt} as the teacher model. 

\paragraph{Compared Methods.}
We compare \ours{} against both specialized multi-hop QA systems and stepwise reasoning baselines.
First, we include TRACE~\citep{fang2024trace}, a KG-chain-based reasoning method, RISE~\citep{he2025rise}, an iterative self-exploration and self-critique framework, and DTKG~\citep{wang2025dtkg}, a KG-verified dual-track reasoning framework.
We also report IRCoT~\citep{trivedi2023interleaving} as a contextual reference for retrieval-based method.
Results for these methods are taken from their original papers.

Second, to isolate the effect of stepwise verification, we compare three reasoning methods for the same generator models:
\begin{itemize}[leftmargin=*, itemsep=0pt, topsep=0pt]
    \item \textit{No Verification}: The generator produces one atomic reasoning step at a time until it reaches a final answer. No verifier is used.
    
    \item \textit{Self-Verification}: The generator evaluates its own intermediate step after each generation and revises the step if it detects an error.
    
    \item \ours{}: The generator is paired with our external verifier. The verifier checks each step against the provided evidence and returns a diagnosis and actionable guidance when the step is invalid.
\end{itemize}

\paragraph{Implementation Details.}
We fine-tune the verifier with QLoRA~\citep{dettmers2023qlora} on the Qwen3-8B non-thinking architecture for 2 epochs. 
We use a learning rate of $1 \times 10^{-4}$ and an effective batch size of 64. 
All experiments are conducted on 4$\times$ A6000 GPUs. 
During inference, we use greedy decoding with temperature 0 for deterministic generation. 
We set the maximum retry attempts $N$ to 3 and the maximum reasoning steps $K$ to 10, which provides the best cost-performance trade-off in Appendix~\ref{sec:appendix_cost}. 
Additional training details are provided in Appendix~\ref{sec:appendix_implementation_details}.

\begin{table*}[t]
\centering
\small
\resizebox{\linewidth}{!}{%
\begin{tabular}{ll ccc ccc ccc ccc c}
\toprule
\multirow{2}{*}{\textbf{Model}} & \multirow{2}{*}{\textbf{Method}} & \multicolumn{3}{c}{\textbf{2Wiki}} & \multicolumn{3}{c}{\textbf{HotpotQA}} & \multicolumn{3}{c}{\textbf{MuSiQue}} & \multicolumn{3}{c}{\textbf{Average}} & \multirow{2}{*}{$\boldsymbol{\Delta}$} \\
\cmidrule(lr){3-5} \cmidrule(lr){6-8} \cmidrule(lr){9-11} \cmidrule(lr){12-14}
& & \textbf{EM} & \textbf{F1} & \textbf{Acc} & \textbf{EM} & \textbf{F1} & \textbf{Acc} & \textbf{EM} & \textbf{F1} & \textbf{Acc} & \textbf{EM} & \textbf{F1} & \textbf{Acc} & \\
\midrule

\multicolumn{15}{l}{\textit{\textbf{Prior Methods}}} \\
\midrule
\multirow{2}{*}{Llama 3 8B} 
 & IRCoT$^\dagger$ & 36.1 & 52.3 & - & 50.8 & 65.7 & - & 27.4 & 36.9 & - & 38.1 & 51.6 & - & - \\
 & TRACE$^\dagger$ & 45.5 & 56.4 & - & 55.1 & 70.0 & - & 33.4 & 40.1 & - & 44.7 & 55.5 & - & - \\
\midrule
\multirow{2}{*}{Llama 3.1 8B} 
 & RISE$^\dagger$ & 32.7 & 43.0 & 49.4 & 30.5 & 42.0 & 40.5 & 11.7 & 22.9 & 21.7 & 25.0 & 36.0 & 37.2 & - \\
 & DTKG$^\dagger$ & - & - & - & 38.2 & - & 85.8 & 18.5 & - & 83.0 & 28.4 & - & 84.4 & - \\
\midrule

\multicolumn{15}{l}{\textit{\textbf{Controlled Comparison}}} \\
\midrule

\multirow{3}{*}{Gemma 3 12B} 
 & No Verification & 50.7 & 65.1 & 79.4 & 51.1 & 68.1 & 86.6 & 41.7 & 56.4 & 67.3 & 47.8 & 63.2 & 77.8 &  \\
 & Self-Verification & 54.3 & 65.0 & 73.3 & 54.6 & 70.5 & 83.4 & 45.6 & 58.1 & 68.0 & 51.5 & 64.5 & 74.9 &  \\
 & \textbf{SAFE (Ours)} & \textbf{72.9} & \textbf{82.7} & \textbf{91.8} & \textbf{59.6} & \textbf{77.5} & \textbf{90.0} & \textbf{53.7} & \textbf{66.6} & \textbf{76.7} & \textbf{62.1} & \textbf{75.6} & \textbf{86.2} & \textbf{+8.4} \\
\midrule

\multirow{3}{*}{Qwen 2.5 14B} 
 & No Verification & 55.4 & 71.0 & 87.2 & 49.5 & 68.4 & 87.9 & 40.4 & 54.1 & 67.1 & 48.4 & 64.5 & 80.7 &  \\
 & Self-Verification & 63.2 & 72.2 & 80.6 & 53.9 & 69.7 & 84.3 & 37.2 & 49.3 & 60.5 & 51.4 & 63.7 & 75.1 &  \\
 & \textbf{SAFE (Ours)} & \textbf{72.8} & \textbf{81.7} & \textbf{91.0} & \textbf{60.8} & \textbf{76.8} & \textbf{91.2} & \textbf{50.7} & \textbf{63.6} & \textbf{74.7} & \textbf{61.4} & \textbf{74.0} & \textbf{85.6} & \textbf{+4.9} \\
\midrule

\multirow{3}{*}{Llama 3.1 8B} 
 & No Verification & 51.0 & 59.6 & 67.5 & 51.1 & 66.7 & 78.9 & 36.3 & 46.9 & 53.4 & 46.1 & 57.7 & 66.6 &  \\
 & Self-Verification & 54.2 & 65.0 & 73.2 & 54.9 & 70.8 & 83.8 & 44.1 & 56.7 & 66.0 & 51.1 & 64.2 & 74.3 &  \\
 & \textbf{SAFE (Ours)} & \textbf{71.7} & \textbf{80.9} & \textbf{90.7} & \textbf{60.4} & \textbf{77.1} & \textbf{90.3} & \textbf{51.8} & \textbf{64.7} & \textbf{75.3} & \textbf{61.3} & \textbf{74.3} & \textbf{85.4} & \textbf{+11.1} \\
\midrule

\multirow{3}{*}{Qwen 3 8B} 
 & No Verification & 53.7 & 66.6 & 85.6 & 47.5 & 65.8 & 87.4 & 40.2 & 53.0 & 68.7 & 47.1 & 61.8 & 80.6 & - \\
 & Self-Verification & 64.0 & 73.4 & 82.9 & 55.0 & 71.2 & 84.4 & 44.5 & 55.7 & 65.2 & 54.5 & 66.8 & 77.5 &  \\
 & \textbf{SAFE (Ours)} & \textbf{71.5} & \textbf{81.3} & \textbf{91.3} & \textbf{61.2} & \textbf{78.3} & \textbf{90.9} & \textbf{53.8} & \textbf{66.7} & \textbf{76.6} & \textbf{62.2} & \textbf{75.4} & \textbf{86.3} & \textbf{+5.7} \\
\midrule

\multirow{3}{*}{Qwen 3 4B} 
 & No Verification & 65.8 & 75.8 & 84.2 & 56.2 & 72.4 & 86.5 & 42.2 & 53.6 & 63.2 & 54.7 & 67.3 & 78.0 & - \\
 & Self-Verification & 65.9 & 74.9 & 82.8 & 56.5 & 70.6 & 82.6 & 38.4 & 48.2 & 56.9 & 53.6 & 64.6 & 74.1 &  \\
 & \textbf{SAFE (Ours)} & \textbf{70.8} & \textbf{79.8} & \textbf{89.4} & \textbf{60.5} & \textbf{75.7} & \textbf{89.0} & \textbf{49.8} & \textbf{61.5} & \textbf{74.5} & \textbf{60.4} & \textbf{72.3} & \textbf{84.3} & \textbf{+6.3} \\ \midrule \midrule

\multirow{3}{*}{\textbf{Total Average}} 
& No Verification & 55.3 & 67.6 & 80.8 & 51.1 & 68.3 & 85.5 & 40.2 & 52.8 & 63.9 & 48.9 & 62.9 & 76.7 &  \\
 & Self-Verification & 60.3 & 70.1 & 78.6 & 55.0 & 70.6 & 83.7 & 42.0 & 53.6 & 63.3 & 52.4 & 64.8 & 75.2 &  \\
 & \textbf{SAFE (Ours)} & \textbf{71.9} & \textbf{81.3} & \textbf{90.8} & \textbf{60.5} & \textbf{77.1} & \textbf{90.3} & \textbf{52.0} & \textbf{64.6} & \textbf{75.6} & \textbf{61.5} & \textbf{74.3} & \textbf{85.6} & \textbf{+8.8} \\  
\bottomrule
\end{tabular}%
}
\caption{
Results on three multi-hop QA benchmarks. 
Evaluation metrics are Exact Match, F1 score, and LLM-as-judge accuracy. 
$\Delta$ denotes the improvement of \ours{} over the best baseline (No Verification or Self-Verification) in terms of average accuracy. 
$\dagger$ indicates results directly reported from the original paper.
}
\label{tab:main_result}
\vspace{-0.5em}
\end{table*}

\subsection{Main Results}
\label{sec:main_results}

Table~\ref{tab:main_result} reports results on the three verified multi-hop QA benchmarks. 
Across datasets and generator models, \ours{} consistently improves answer accuracy over both no-verification and self-verification baselines. 
On average, \ours{} achieves an 8.8 pp accuracy gain over the strongest baseline and improves EM by 9.1 pp, from 52.4\% to 61.5\%.

The comparison with no-verification shows that atomic step-wise generation is a strong starting point, but it is not sufficient. 
Even when the generator is prompted to produce one atomic step at a time, unsupported entities, invalid relations, and skipped intermediate steps can still appear. 
\ours{} addresses these failures by checking each intermediate step against the evidence before the reasoning process continues.

The comparison with self-verification highlights the limitation of relying on the generator to critique itself. 
Self-verification decreases average accuracy from 76.7\% to 75.2\%, and the drop is especially large for Qwen2.5-14B, where accuracy falls from 80.7\% to 75.1\%. 
This supports the need for an external verifier, since the same model that generates a reasoning step may fail to locate its own grounding errors or may reinforce an invalid trajectory when asked to verify it.

\ours{} is especially effective on MuSiQue, the most challenging benchmark in our evaluation due to its longer and more compositional reasoning paths. 
On MuSiQue, \ours{} improves average accuracy by 11.7 pp, from 63.9\% to 75.6\%, and improves EM by 10.0 pp. 
These gains suggest that stepwise verification is most beneficial when errors can accumulate across multiple reasoning steps.

Finally, the results show that \ours{} is not tied to a single generator architecture. 
Despite differences in model family and scale, generators paired with \ours{} reach a similar high-performance range of roughly 84\%--86\% average accuracy. 
This suggests that the gains come from external stepwise verification rather than from the characteristics of a particular generator.
Overall, the main results show that evidence-grounded multi-hop QA benefits from an LLM-as-verifier framework that checks and corrects intermediate reasoning steps during generation.

To assess robustness to the evaluation model, we additionally evaluate all outputs with Llama-3.3-70B-Instruct~\cite{grattafiori2024llama}, which is independent of the model used for verifier supervision. SAFE retains the best average accuracy across all five generators, consistent with our main
evaluation (Appendix Table~\ref{tab:llama_judge}).

\section{Analysis: What Makes \ours{} Effective?}
\label{sec:analysis}
The main results show that \ours{} improves answer accuracy across datasets and generator models. 
We analyze the sources and robustness of these gains through five findings, covering structured step-level feedback, train-time supervision, teacher strength, generator strength, and incomplete-evidence scenarios. 
Additional efficiency and failure-pattern analyses are provided in Appendix~\ref{sec:appendix_cost}.

\paragraph{Finding 1: Structured step-level feedback matters.}

\begin{table}[t]
\centering
\small
\resizebox{\columnwidth}{!}{
\begin{tabular}{llcccc}
\toprule
\textbf{Model} & \textbf{Setting} & \textbf{2Wiki} & \textbf{Hotpot} & \textbf{MSQ} & \textbf{Avg} \\
\midrule
\multirow{3}{*}{Gemma 12B}
 & Diagnosis only & 86.5 & 88.0 & 71.2 & 81.9 \\
 & Guidance only  & 90.9 & 88.8 & 76.9 & 85.5 \\
 & Full           & 91.8 & 90.0 & 76.7 & 86.2 \\
\midrule
\multirow{3}{*}{Llama 8B}
 & Diagnosis only & 83.7 & 83.6 & 66.5 & 77.9 \\
 & Guidance only  & 89.5 & 89.5 & 77.2 & 85.4 \\
 & Full           & 90.7 & 90.3 & 75.3 & 85.4 \\
\midrule
\multirow{3}{*}{Qwen 8B}
 & Diagnosis only & 90.1 & 88.6 & 74.1 & 84.3 \\
 & Guidance only  & 91.2 & 90.0 & 76.6 & 85.9 \\
 & Full           & 91.3 & 90.9 & 76.6 & 86.3 \\
\bottomrule
\end{tabular}
}
\caption{
Ablation of verifier feedback formats. 
Guidance-only feedback provides the strongest signal, while the full format combines diagnosis and guidance for more stable correction.
}
\label{tab:diagnosis_guidance}
\end{table}

We first analyze whether \ours{} improves performance through generic feedback or structured step-level feedback. 
To isolate this effect, we ablate the verifier's response format with three settings: \textit{Diagnosis only}, \textit{Guidance only}, and \textit{Full}. 
Diagnosis-only explains the error without providing correction guidance, guidance-only provides correction guidance without explicit diagnosis, and full uses both.

Table~\ref{tab:diagnosis_guidance} shows that guidance provides the strongest signal. 
Across Gemma-3-12B, Qwen3-8B, and Llama-3.1-8B, guidance-only substantially outperforms diagnosis-only, indicating that simply explaining what went wrong is insufficient. 
The verifier must also provide information that helps the generator revise the invalid step. 
At the same time, the full verifier is the most stable overall, suggesting that diagnosis contributes to a reliable feedback interface. 
Thus, \ours{} goes beyond generic feedback by using step-level verification to support correction during generation.

\begin{table}[t]
\centering
\small
\resizebox{\linewidth}{!}{%
\begin{tabular}{lccc}
\toprule
\textbf{Drop Category} & \textbf{Gemma 3 12B} & \textbf{Llama 3.1 8B} & \textbf{Qwen 3 8B} \\
\midrule

No Drop 
& 86.2 
& 85.4 
& 86.3 \\

Procedural 
& 85.2 \textcolor{red}{\scriptsize (-1.0)} 
& 82.9 \textcolor{red}{\scriptsize (-2.5)} 
& 84.4 \textcolor{red}{\scriptsize (-1.9)} \\

Attribution 
& 84.4 \textcolor{red}{\scriptsize (-1.8)} 
& 82.8 \textcolor{red}{\scriptsize (-2.6)} 
& 84.5 \textcolor{red}{\scriptsize (-1.8)} \\

Logical 
& 85.5 \textcolor{red}{\scriptsize (-0.7)} 
& 84.3 \textcolor{red}{\scriptsize (-1.1)} 
& 85.0 \textcolor{red}{\scriptsize (-1.3)} \\

\bottomrule
\end{tabular}%
}
\caption{
Performance degradation (average score) under different error category removals. Values in parentheses indicate absolute change ($\Delta$) from the No Drop setting.
}
\label{tab:drop_error_types}
\vspace{-1em}
\end{table}

We further test whether the error categories are useful for verifier training by removing each major category. 
As shown in Table~\ref{tab:drop_error_types}, dropping any category degrades performance across generator models. 
Removing Procedural and Attribution categories is especially harmful, indicating that the verifier benefits from distinguishing reasoning-path validity from evidence-grounding failures. 
These results show that the feedback categories provide useful supervision for detecting and correcting invalid reasoning steps.

\paragraph{Finding 2: Train-time verification improves verifier supervision.}

\begin{table}[t]
\centering
\small
\resizebox{\linewidth}{!}{%
\begin{tabular}{llcccc}
\toprule
\textbf{Model} & \textbf{Training Data} & \textbf{2Wiki} & \textbf{Hotpot} & \textbf{MSQ} & \textbf{Avg} \\
\midrule

\multirow{2}{*}{Gemma-3-12B}
 & Verified & 91.8 & 90.0 & 76.7 & 86.2 \\
 & + Unverifiable  & 88.5 \textcolor{red}{\scriptsize (-3.3)} & 89.7 \textcolor{red}{\scriptsize (-0.3)} & 74.9 \textcolor{red}{\scriptsize (-1.8)} & 84.4 \textcolor{red}{\scriptsize (-1.8)} \\
\midrule

\multirow{2}{*}{Qwen2.5-14B}
 & Verified & 90.2 & 90.1 & 76.6 & 85.6 \\
 & + Unverifiable  & 88.0 \textcolor{red}{\scriptsize (-2.2)} & 87.9 \textcolor{red}{\scriptsize (-2.2)} & 70.9 \textcolor{red}{\scriptsize (-5.7)} & 82.3 \textcolor{red}{\scriptsize (-3.3)} \\
\midrule

\multirow{2}{*}{Llama-3.1-8B}
 & Verified & 90.7 & 90.3 & 75.3 & 85.4 \\
 & + Unverifiable  & 88.2 \textcolor{red}{\scriptsize (-2.5)} & 87.5 \textcolor{red}{\scriptsize (-2.8)} & 72.3 \textcolor{red}{\scriptsize (-3.0)} & 82.7 \textcolor{red}{\scriptsize (-2.7)} \\
\midrule

\multirow{2}{*}{Qwen3-8B}
 & Verified & 91.3 & 90.9 & 76.6 & 86.3 \\
 & + Unverifiable  & 88.0 \textcolor{red}{\scriptsize (-3.3)} & 89.5 \textcolor{red}{\scriptsize (-1.4)} & 75.0 \textcolor{red}{\scriptsize (-1.6)} & 84.2 \textcolor{red}{\scriptsize (-2.1)} \\
\midrule

\multirow{2}{*}{Qwen3-4B}
 & Verified & 89.4 & 89.0 & 74.5 & 84.3 \\
 & + Unverifiable  & 86.6 \textcolor{red}{\scriptsize (-2.8)} & 86.8 \textcolor{red}{\scriptsize (-2.2)} & 70.6 \textcolor{red}{\scriptsize (-3.9)} & 81.3 \textcolor{red}{\scriptsize (-3.0)} \\
\midrule\midrule

\multirow{2}{*}{Total}
 & Verified & 90.7 & 90.1 & 75.9 & 85.6 \\
 & + Unverifiable  & 87.9 \textcolor{red}{\scriptsize (-2.8)} & 88.3 \textcolor{red}{\scriptsize (-1.8)} & 72.7 \textcolor{red}{\scriptsize (-3.2)} & 83.0 \textcolor{red}{\scriptsize (-2.6)} \\
\bottomrule
\end{tabular}%
}
\caption{
Effect of adding unverifiable instances to verifier training data. 
Training with unverifiable supervision consistently degrades accuracy, showing the importance of train-time benchmark verification.
}
\label{tab:train_with_noise}
\end{table}

We next examine how train-time verification affects verifier training. 
As shown in Table~\ref{tab:train_with_noise}, we compare our standard verifier, trained on the verified training set, with a variant trained on the same data plus 5.3k instances flagged as unverifiable by our KG-grounded verification pipeline. 
These additional instances follow the same data generation format but do not satisfy the evidence-grounding constraints.

Adding unverifiable instances consistently hurts performance, causing an average accuracy drop of 2.6 pp. 
This indicates that unverifiable supervision introduces contradictory training signals, causing the verifier to approve steps that cannot be supported by the provided evidence. 
This shows that train-time verification helps construct reliable supervision for the verifier.

We also evaluate whether the iterative refinement strategy in Section~\ref{sec:method_train_time} improves the verifier beyond synthetic error injection. 
As shown in Appendix Table~\ref{tab:synthetic_vs_refined}, targeted refinement improves average accuracy by 2.9 pp, with the largest gain on MuSiQue (+4.2 pp). 
This suggests that synthetic perturbations provide broad coverage of error categories, while targeted refinement helps the verifier learn model-specific failures that occur during actual inference.

\paragraph{Finding 3: \ours{} is not merely strong teacher distillation.}

\begin{table}[t]
    \centering
    \resizebox{\columnwidth}{!}{
        \begin{tabular}{lcccc}
            \toprule
            Model & No Verification & Self-Verification & \ours{} (20B) & \ours{} (120B) \\
            \midrule
            Gemma3 12B    & 78.0 & 73.0 & 84.7 & 89.7 \\
            Qwen3 8B      & 83.7 & 76.3 & 86.7 & 89.0 \\
            Llama 3.1 8B  & 66.3 & 56.7 & 78.7 & 83.0 \\
            \bottomrule
        \end{tabular}
    }
    \caption{Performance with weaker teacher model's verification signal (gpt-oss-20b). Results are averaged over 100 examples per benchmark.}
    \label{tab:weaker_teacher}
\end{table}

Because gpt-oss-120b is used to construct verifier training data, a natural question is whether \ours{}'s gains mainly come from the strength of this teacher model. 
To test this, we replace gpt-oss-120b with the weaker gpt-oss-20b for generating verifier supervision while keeping the rest of the pipeline unchanged.

Table~\ref{tab:weaker_teacher} shows that \ours{} remains effective even with the weaker teacher. 
For Gemma-3-12B, Qwen3-8B, and Llama-3.1-8B, the verifier trained with gpt-oss-20b supervision still outperforms both no-verification and self-verification baselines. 
The stronger gpt-oss-120b teacher provides additional gains, but the consistent improvement with gpt-oss-20b indicates that the benefit is not simply inherited from a strong teacher. 
Moreover, the teacher is used only offline for data construction. At inference-time, \ours{} uses a separately trained verifier.

\paragraph{Finding 4: \ours{} remains effective for stronger generators.}

\begin{table}[t]
    \centering
    \small
    \resizebox{\columnwidth}{!}{
        \begin{tabular}{llccccc}
            \toprule
            Model & Method & 2Wiki & Hotpot & MSQ & Avg & $\Delta$ \\
            \midrule
            \multirow{3}{*}{Gemini 3.1 FL} 
                & No Verification   & 94.0 & 90.0 & 71.0 & 85.0 & -- \\
                & Self-Verification & 91.0 & 86.0 & 71.0 & 82.7 & -- \\
                & \ours{}           & 94.0 & 94.0 & 76.0 & 88.0 & +3.0 \\
            \midrule
            \multirow{3}{*}{GPT-5.4 Mini} 
                & No Verification   & 85.0 & 93.0 & 82.0 & 86.7 & -- \\
                & Self-Verification & 92.0 & 87.0 & 74.0 & 84.3 & -- \\
                & \ours{}           & 97.0 & 93.0 & 80.0 & 90.0 & +3.3 \\
            \bottomrule
        \end{tabular}
    }
    \caption{
    Subset evaluation on stronger reasoning models. 
    Results are averaged over 100 examples per benchmark.
    }
    \label{tab:stronger_generator}
\end{table}

To test whether \ours{}'s external verification remains useful for stronger generators beyond the small to mid-sized models in our main experiments, we additionally evaluate Gemini 3.1 Flash-Lite and GPT-5.4 mini.
Table~\ref{tab:stronger_generator} shows that \ours{} still improves over the strongest baseline. 
For Gemini 3.1 Flash-Lite, \ours{} improves average accuracy from 85.0\% to 88.0\%, a gain of 3.0 pp. 
For GPT-5.4 mini, \ours{} improves average accuracy from 86.7\% to 90.0\%, a gain of 3.3 pp. 
These results suggest that stronger reasoning ability does not remove the benefit of explicit evidence-grounded verification.

\paragraph{Finding 5: \ours{} extends to incomplete-evidence settings.}

\begin{table}[t]
    \centering
    \resizebox{\columnwidth}{!}{
        \begin{tabular}{llcc}
            \toprule
            Model & Dataset & W/O Retrieval & W/ Retrieval \\
            \midrule
            \multirow{3}{*}{Gemma 3 12B} & 2Wiki & 44 & 92 \\
             & HotpotQA & 55 & 90 \\
             & MuSiQue & 43 & 81 \\
            \midrule
            \multirow{3}{*}{Qwen 3 8B} & 2Wiki & 31 & 95 \\
             & HotpotQA & 57 & 88 \\
             & MuSiQue & 42 & 77 \\
            \midrule
            \multirow{3}{*}{Llama 3.1 8B} & 2Wiki & 35 & 93 \\
             & HotpotQA & 53 & 87 \\
             & MuSiQue & 44 & 69 \\
            \midrule
            \textbf{Total Avg.} & & \textbf{44.9} & \textbf{85.8 (+40.9)} \\
            \bottomrule
        \end{tabular}
    }
    \caption{Performance of with/without retrieval under incomplete evidence setting. Results are averaged over 100 examples per benchmark.}
    \label{tab:retrieval-performance}
\end{table}

Our main evaluation assumes that the required evidence is fully included in the provided context. 
To examine whether \ours{} can extend beyond this setting, we conduct an incomplete-evidence experiment where one gold passage is removed from the initial context. 
We add \textit{Missing Evidence} as an additional error category and train the verifier to produce a search query when the current context is insufficient.

At inference-time, we compare answering with the incomplete context against retrieving the top-3 passages from the full corpus using verifier-generated search queries and BM25~\citep{robertson2009probabilistic}. 
Table~\ref{tab:retrieval-performance} shows that using the retrieved passages improves average accuracy from 44.9\% to 85.8\%, yielding a gain of 40.9 pp. 
While this experiment does not aim to solve full open-domain retrieval, it demonstrates that \ours{} can identify missing-evidence failures and use them to guide retrieval.

\section{Conclusion}
We introduced \ours{}, an LLM-as-verifier framework for evidence-grounded multi-hop QA. 
Rather than treating verification as post-hoc judgment, \ours{} verifies intermediate reasoning steps during generation. 
By representing reasoning steps as evidence-grounded atomic units, \ours{} constructs reliable verifier supervision at train-time and provides feedback for correcting invalid reasoning at inference-time.
Across three multi-hop QA benchmarks, \ours{} achieves an average +8.8 pp accuracy gain over baselines. 
Our analyses show that these gains arise from structured step-level verification and feedback, highlighting the importance of checking and correcting intermediate reasoning steps in evidence-grounded multi-hop QA.\label{sec:conclusion}

\section*{Acknowledgement}

This work was supported by Institute of Information \& communications Technology Planning \& Evaluation (IITP) grant funded by the Korea government (MSIT) (No. RS-2026-25522672, Development of Unified Reasoning Technology Mimicking Human Cognition for Hierarchical Understanding and Unbounded Problem Solving) and Institute of Information \& communications Technology Planning \& Evaluation (IITP) grant funded by the Korea government(MSIT) [NO.RS-2021-II211343, Artificial Intelligence Graduate School Program (Seoul National University)].

\section*{Limitations}

While we utilized a KG to ensure the quality of our verification pipeline, we acknowledge the potential for mislabeling due to inherent noise or incomplete coverage in the KG. To address this, rather than aggressively excluding instances that might contain valuable reasoning trajectories, we provide the original dataset with our feedback labels as an augmented resource. This non-destructive approach allows future researchers to audit, refine, or leverage the full context of the data while accounting for potential label noise in automated diagnostics.

Furthermore, the efficacy of the \ours{} framework inherently relies on the reasoning generator's instruction-following capabilities. We observed instances where, despite the feedback model successfully producing accurate diagnoses and explicit corrective guidance, the reasoning generator failed to effectively incorporate these instructions into its subsequent generation step. Such instruction-following failures can lead to uncorrected errors or repeated mistakes within the iterative loop. This highlights a dependency on the base language model's capacity to adhere to complex, stepwise constraints, suggesting that future work must address the alignment between corrective feedback and the generator's execution.

\section*{Ethics Statement}

This work introduces \ours{}, a framework designed to enhance the reliability of LLMs in evidence-grounded multi-hop reasoning through LLM-as-verifier framework. We acknowledge several ethical considerations and potential broader impacts associated with its development and deployment.

\paragraph{Transparency and Interpretability}
A persistent ethical challenge with LLMs is their ``black-box'' nature, which obscures the rationale behind generated answers. \ours{} addresses this opacity by decomposing multi-hop reasoning into atomic, evidence-grounded steps that can be independently verified. This design makes the reasoning process more transparent by allowing researchers and practitioners to inspect which intermediate steps are supported, unsupported, or corrected. By exposing step-level diagnoses and actionable guidance, \ours{} can support more precise error analysis, auditing, and accountability in AI-assisted reasoning systems.

\paragraph{Computational Efficiency and Environmental Impact}
Iterative reasoning and correction frameworks can incur additional computational overhead due to repeated generation and verification steps. \ours{} mitigates this issue by using efficient KV caching mechanisms to reduce redundant computation during iterative generation. While \ours{} still introduces additional inference-time verification cost compared to single-pass generation, our cost analysis shows that it can improve multi-hop reasoning reliability with a moderate computational overhead. This aligns with the broader need for more efficient and sustainable AI systems.

\paragraph{Bias and Fairness}
The performance and behavior of \ours{} depend on the underlying pre-trained LLMs and the external knowledge sources used for reasoning, such as retrieved Wikipedia passages. These models and corpora may contain historical, cultural, or demographic biases. While \ours{} targets procedural, attribution, logical, and answer-level reasoning errors, it is not explicitly designed to detect or mitigate social biases. Users should therefore remain cautious, as the system may reproduce or amplify biases present in the source material.

\paragraph{Licenses and Terms of Use.}
We use publicly available multi-hop QA benchmarks and open-source language models in accordance with their respective licenses and terms of use. We cite the original creators of all datasets, models, and software artifacts used in our experiments. Our use of these artifacts is limited to research and evaluation purposes.

\paragraph{AI Usage}
AI-based tools were used for grammar correction and language refinement. The authors take full responsibility for all scientific content, including the methodology, experiments, and conclusions.

\newpage
\bibliography{custom}

\clearpage
\appendix

\section{Appendix}

\subsection{Overview}

This appendix provides implementation details, supplementary analyses, and qualitative examples that support the findings in the main text. The contents are organized as follows:

\begin{itemize}[leftmargin=*]
    \item Section~\ref{sec:appendix_implementation_details} provides implementation details for training the verifier, including hyperparameters, LoRA settings, decoding configurations, and hardware specifications.
    
    \item Section~\ref{sec:appendix_kg_filtering} presents qualitative examples from our KG-grounded benchmark verification, including false positives and unverifiable benchmark instances flagged by the pipeline (see Table~\ref{tab:false_positive_example}, Table~\ref{tab:noise_examples}, and Figure~\ref{fig:multihop_error_examples}).
    
    \item Section~\ref{sec:appendix_evaluation_datasets} describes the preprocessing and sampling procedure for constructing the verified evaluation pool used in our multi-hop QA experiments.
    
    \item Section~\ref{sec:appendix_training_dataset} provides details of the verifier training data, including examples, error-step positions, and dataset statistics (see Figure~\ref{fig:training_data_example}, Figure~\ref{fig:error_position_statistics}, and Table~\ref{tab:training_data_unique_question}).
    
    \item Section~\ref{sec:appendix_case_study} presents qualitative case studies comparing \ours{} with the self-verification baseline. These examples illustrate how the external verifier handles common multi-hop failure modes, including lexical distractors, premature attribution, and self-verification collapse (see Table~\ref{tab:qualitative_comparison}, Table~\ref{tab:premature_attribution_appendix}, and Table~\ref{tab:self_feedback_collapse}).
    
    \item Section~\ref{sec:appendix_synthetic_only_vs_full} reports an ablation study on targeted refinement, comparing a verifier trained only on synthetic errors with the final verifier trained with additional hard cases (see Table~\ref{tab:synthetic_vs_refined}).
    
    \item Section~\ref{sec:appendix_cost} provides detailed efficiency and failure-pattern analyses, including the cost-accuracy comparison between \ours{} and self-verification, the sensitivity of performance and cost to the maximum reasoning steps $K$ and retry limit $N$, the use of Prefix KV caching, and the distributions of detected error types and error positions.
    
    \item Section~\ref{sec:appendix_prompts} lists the prompts used throughout our pipeline, including reasoning generation, stepwise verification, diagnostic error injection, and KG-grounded benchmark verification (see Figure~\ref{fig:plan_generation_prompt_2wiki}--\ref{fig:logical_path_discovery_prompt}).
\end{itemize}

\begin{table*}[t]
\centering
\renewcommand{\arraystretch}{1.2} 
\resizebox{\linewidth}{!}{
\begin{tabular}{@{} >{\raggedright\arraybackslash}m{0.2\linewidth} >{\raggedright\arraybackslash}m{0.21\linewidth} >{\raggedright\arraybackslash}m{0.59\linewidth} @{}}
\toprule
\textbf{Category} & \textbf{Error Type} & \textbf{Definition} \\ \midrule
\multirow{4}{*}{\parbox{\linewidth}{
\textbf{Procedural} \\
{\footnotesize \textit{(Attribution \&\\Logical Steps)}}
}
} & Overthinking$^\dagger$ & Engaging in excessive extraction or derivation steps even after the target node ($e_{ans}$) has been reached. \\
 & Inefficiency$^\dagger$ & Generating textual steps that do not correspond to any valid KG operation (i.e., no node or edge traversal). \\
 & Off-topic$^\dagger$ & Extracting entities or traversing paths that form a disconnected subgraph, not contributing to the target path. \\
 & Redundancy$^\dagger$ & Repeating the extraction of a node $e$ or traversal of an edge $(e_a, r, e_b)$ already documented in previous steps. \\ \midrule
\multirow{4}{*}{\parbox{\linewidth}{
\textbf{Attribution} \\
{\footnotesize \textit{(Attribution Steps)}}
}
} & Unsupported & Incorporating an entity $e$ or relationship $r$ that is completely absent from the evidence subgraph $\mathcal{G}$. \\
 & Premature Attribution & Attempting to connect a valid entity $e_n$ to $e_m$ while bypassing the necessary intermediate edge $(e_m, r, e_n)$. \\
 & Information Miss & Overlooking a necessary node $e$ that is explicitly present and available within $\mathcal{G}$. \\
 & Contradictory & Generating a node or edge relation that directly conflicts with explicit facts within $\mathcal{G}$. \\\midrule
{\parbox{\linewidth}{
\textbf{Logical} \\
{\footnotesize \textit{(Logical Steps)}}
}
} & Logical Fallacy & Inferring a new relation $r_{inferred}$ that cannot be logically proven via established KG traversal rules over extracted paths. \\ \midrule
{\parbox{\linewidth}{
\textbf{Final Answer} \\
{\footnotesize \textit{(Final Answer Steps)}}
}
} & Wrong Conclusion & Producing a target $e_{ans}$ that is missing from the reasoning path or fails to logically terminate the grounded sequence. \\ \bottomrule
\end{tabular}%
}
\caption{
Feedback categories used by \ours{} for stepwise verification.
Each invalid reasoning step is assigned one mutually exclusive error type, which determines the feedback provided for correction.
Procedural errors marked with $\dagger$ are checked first because they concern the structure and progress of the reasoning path before evidence attribution and logical consistency are assessed.
}
\vspace{-1em}
\label{tab:error_taxonomy}
\end{table*}

\subsection{Verifier Training Details}
\label{sec:appendix_implementation_details}

As described in Section~\ref{sec:experimental_setup}, we fine-tune the verifier using Parameter-Efficient Fine-Tuning (PEFT) with QLoRA~\citep{dettmers2023qlora}. 
We set the LoRA rank to $r=64$, the scaling factor to $\alpha=128$, and the dropout rate to $0.05$. 
LoRA is applied to all linear modules, including the attention projections ($q$, $k$, $v$, and $o\_proj$) and the MLP layers ($gate$, $up$, and $down\_proj$). 
For optimization, we use the AdamW optimizer~\citep{loshchilov2017decoupled} with a cosine learning-rate scheduler and a warmup ratio of $0.03$.

\subsection{KG-Grounded Benchmark Verification: Examples}
\label{sec:appendix_kg_filtering}

\begin{figure*}[t]
    \centering
    \includegraphics[width=0.7\textwidth]{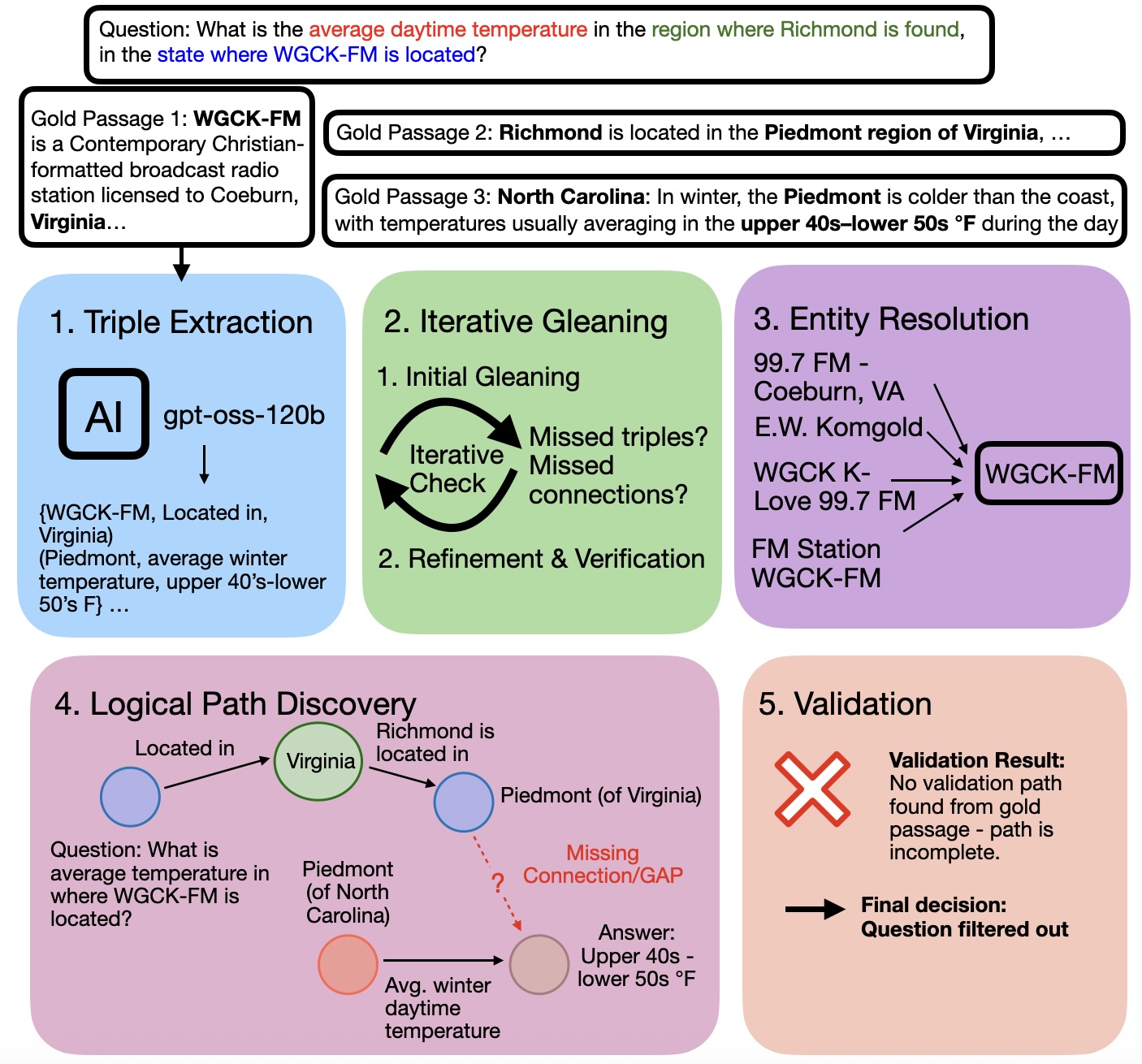}
    \caption{
Overview of our KG-grounded benchmark verification pipeline. 
The example shows an unverifiable reasoning path, where the fact is supported by a passage about \textit{North Carolina's Piedmont} rather than \textit{Virginia's Piedmont}. 
Our pipeline detects such evidence-grounding gaps and removes the instance from the verified set.
}
    \label{fig:kg_filtering_pipeline}
\end{figure*}

\paragraph{Dataset.}
We publicly release \textbf{SAFE-Verified-MultiHopQA},
which includes evidence-verified and filtered versions
of 2WikiMultiHopQA, HotpotQA, and MuSiQue.
The datasets are available at
\url{https://huggingface.co/datasets/Daeyongkwon98/SAFE-Verified-MultiHopQA}.

\paragraph{False Positive Examples.}

Table~\ref{tab:false_positive_example} presents qualitative examples of false positives from our KG-grounded benchmark verification pipeline. 
Here, a false positive refers to a valid multi-hop query that the pipeline incorrectly flags as unverifiable.

As illustrated by the three representative cases, the provided gold passages contain the necessary evidence to complete the intended reasoning paths. 
For example, the queries require comparing directors' ages, matching a specific birth date to a TV show judge, or tracing a newspaper's sponsorship to its base city. 
In these cases, the textual evidence is clear, self-contained, and sufficient to support an unambiguous answer.

These examples show that misclassifications can occur even for standard, well-structured reasoning paths. 
They are not limited to unusually complex or poorly phrased queries. 
This observation is important because it suggests that our verification pipeline does not simply remove difficult examples or penalize complex reasoning structures. 
The verified set therefore preserves challenging multi-hop reasoning instances while removing cases that are not verifiable under our evidence-grounding protocol.

\begin{table*}[t!]
\centering
\small

\begin{tabularx}{\textwidth}{@{}p{0.25\textwidth} p{0.31\textwidth} p{0.13\textwidth} X@{}}
\toprule
\textbf{Question} & \textbf{Gold Passages (Truncated)} & \textbf{Answer} & \textbf{Reasoning Analysis} \\ \midrule

Which film has the director who is older, Koeputkiaikuinen Ja Simon Enkelit or Indiana Jones And The Temple Of Doom? & 
[1] \textit{Koeputkiaikuinen... directed, written and starring Spede Pasanen...} \newline 
[2] \textit{Indiana Jones and the Temple of Doom... directed by Steven Spielberg...} \newline
[3] \textit{Spede Pasanen (10 April 1930 – 7 Sept 2001) was a Finnish film director...} \newline
[4] \textit{Steven Spielberg (born December 18, 1946) is an American filmmaker...} & 
Koeputkiaikuinen Ja Simon Enkelit & 
The passages provide all necessary components for the multi-hop chain: (1) identifies Pasanen and Spielberg as respective directors, and (2) provides their birth years (1930 vs. 1946), enabling a correct age comparison. \\ \midrule

What judge who was on the show ``Dancing with the Stars'' was born on the 31st of March, 1963? & 
[1] \textit{Paul Joseph Mercurio (born 31 March 1963) is an Australian actor, dancer, and TV presenter...} \newline 
[2] \textit{...with the exception of Paul Mercurio who wanted to focus more on the Australian version of ``Dancing with the Stars''.} & 
Paul Joseph Mercurio & 
The passages provides a clear multi-hop chain: (1) identifies the specific birth date for Paul Joseph Mercurio, and (2) confirms his role as a judge on the mentioned TV show, allowing for an unambiguous match. \\ \midrule

Beena Sarwar is the editor of a peace initiative sponsored by a newspaper based in what city? & 
[1] \textit{Beena Sarwar is the Pakistan Editor of the Aman ki Asha initiative... jointly sponsored by the Jang group...} \newline 
[2] \textit{The Daily Jang is an Urdu newspaper based in Karachi, Pakistan.} & 
Karachi, Pakistan & 
The passages provide a complete, self-contained chain: (1) connects Beena Sarwar to the Aman ki Asha initiative, (2) identifies Jang group as the sponsor, and (3) explicitly states the newspaper's base city as Karachi. \\ \bottomrule

\end{tabularx}
\caption{
False positive examples from our KG-grounded benchmark verification pipeline. 
These valid instances are incorrectly flagged as unverifiable, showing that misclassifications can occur on standard, well-structured reasoning paths rather than only on unusually complex queries. 
This suggests that the pipeline does not simply remove examples based on reasoning difficulty.
}
\label{tab:false_positive_example}

\vspace{5em}

\begin{tabularx}{\textwidth}{@{}p{0.25\textwidth} p{0.33\textwidth} p{0.1\textwidth} X@{}}
\toprule
\textbf{Question} & \textbf{Gold Passages (Truncated)} & \textbf{Answer} & \textbf{Error Description} \\ \midrule

Where was the place of death of the composer of film The Private Lives Of Elizabeth And Essex? & 
[1] \textit{...The score was composed by Erich Wolfgang Korngold...} \newline 
[2] \textit{...Although his late classical Romantic compositions were no longer as popular when he died in 1957...} & 
Hollywood & 
\textbf{Missing Evidence:} The passages state that the composer (Korngold) died in 1957, but explicitly lack information regarding his place of death. \\ \midrule

Where did the creator of the Allegory of Isabella d'Este's Coronation die? & 
[1] The Allegory... is a painting by the Italian Renaissance painter Lorenzo Costa the Elder... \newline 
[2] \textit{Lorenzo Costa the Younger (1537–1583) was an Italian painter... active in his native city of Mantua.} & 
Mantua & 
\textbf{Entity Disambiguation Failure:} The dataset incorrectly matches ``Lorenzo Costa the Elder'' (the creator) with his grandson ``Lorenzo Costa the Younger,'' conflating two distinct entities. \\ \midrule

What movie did Lauren Bacall's spouse win his only Oscar? & 
[1] \textit{Humphrey Bogart... received three Academy Award nominations for Best Actor, winning one (for The African Queen).} \newline 
[2] \textit{Bold Venture was a syndicated radio series starring Humphrey Bogart and Lauren Bacall...} & 
The African Queen & 
\textbf{External Knowledge Requirement:} The passages establish that Bogart and Bacall co-starred in a radio series, but omit their marital status. Answering requires parametric external knowledge. \\ \midrule

What nationality is the director of film 3000 Nights? & 
[1] \textit{3000 Nights is a 2015 drama film directed by Mai Masri...} \newline 
[2] \textit{Mai Masri... is a Palestinian filmmaker, director and producer.} & 
Lebanon & 
\textbf{Wrong Answer:} The passages explicitly identify the director as ``Palestinian,'' yet the ground-truth answer is ``Lebanon,'' which contradicts the provided context. \\ \bottomrule

\end{tabularx}
\caption{
Representative examples of benchmark instances flagged as unverifiable by our KG-grounded benchmark verification pipeline. 
These cases include evidence-grounding failures, such as missing necessary information in the gold passages or ground-truth answers that contradict the provided evidence.
}
\label{tab:noise_examples}
\vspace{-1em}
\end{table*}

\paragraph{Unverifiable Benchmark Examples.}

\begin{figure}[t]
{
\centering
    \includegraphics[width=\linewidth]{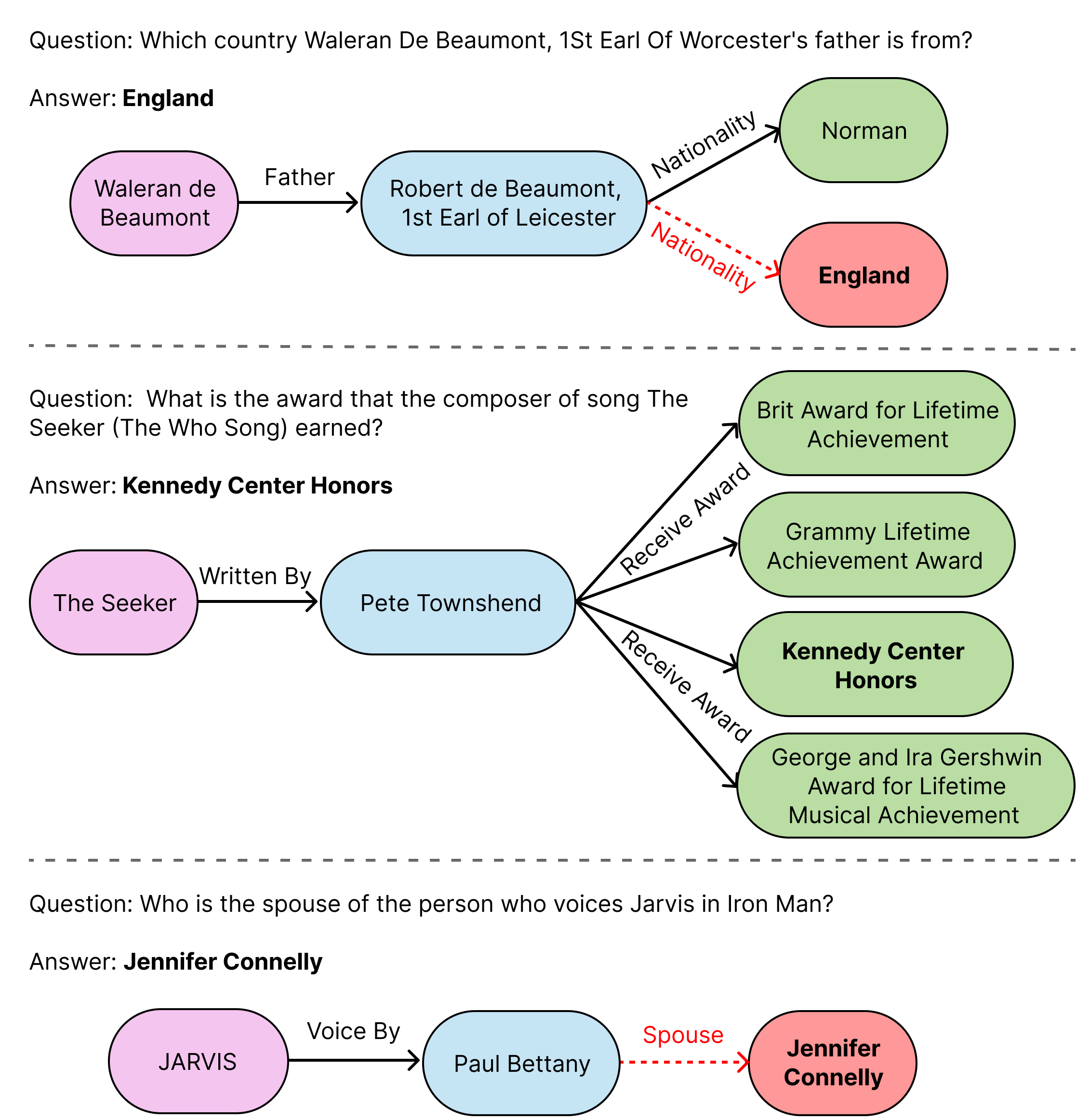}
    \caption{
Representative benchmark instances flagged as unverifiable by our KG-grounded benchmark verification pipeline. 
The examples show an incorrect ground-truth answer that contradicts the provided evidence (top), an ambiguous question without a unique answer (middle), and insufficient context where the gold passages do not support a complete evidence-grounded reasoning path (bottom).
}
    \label{fig:multihop_error_examples}
}
\end{figure}

In contrast to the false positives above, Table~\ref{tab:noise_examples} and Figure~\ref{fig:multihop_error_examples} illustrate benchmark instances that are genuinely unverifiable under our KG-grounded protocol. 
These examples highlight why train-time benchmark verification is necessary for constructing reliable verifier supervision and evaluation data.

First, some instances contain insufficient context, as shown in Table~\ref{tab:noise_examples} and at the bottom of Figure~\ref{fig:multihop_error_examples}. 
In these cases, the provided gold passages lack the evidence needed to complete the reasoning path. 
For example, a passage may specify a composer's year of death but omit the location required by the question.

Other instances suffer from entity confusion. 
Table~\ref{tab:noise_examples} shows a case where two distinct historical figures, Lorenzo Costa the Elder and Lorenzo Costa the Younger, are conflated to form an invalid reasoning path. 
Some examples also contain incorrect ground-truth labels that contradict the provided text, such as identifying a director's nationality as Lebanese when the gold passage states that she is Palestinian, as shown at the top of Figure~\ref{fig:multihop_error_examples}. 
In addition, the middle example in Figure~\ref{fig:multihop_error_examples} shows an ambiguous query that does not support a single definitive answer.

By removing insufficient, contradictory, and ambiguous instances from the verified set, our pipeline ensures that models are evaluated on evidence-grounded multi-hop reasoning rather than on guessing, hallucination, or parametric memorization.

\subsection{Evaluation Dataset Preprocessing Details}
\label{sec:appendix_evaluation_datasets}

For evaluation, we use a constrained distractor setting with exactly 10 passages per question across all benchmarks. 
For 2WikiMultihopQA and HotpotQA, we use the 10 passages provided in their standard distractor settings. 
For MuSiQue, which originally provides 20 passages per instance, we retain all gold supporting passages and randomly sample the remaining passages from the distractor set to satisfy the 10-passage constraint. 
Finally, we randomly shuffle the passage order for every instance in all three datasets to prevent models from exploiting positional shortcuts.

\begin{figure}[t!]
{
\centering
    \includegraphics[width=\linewidth]{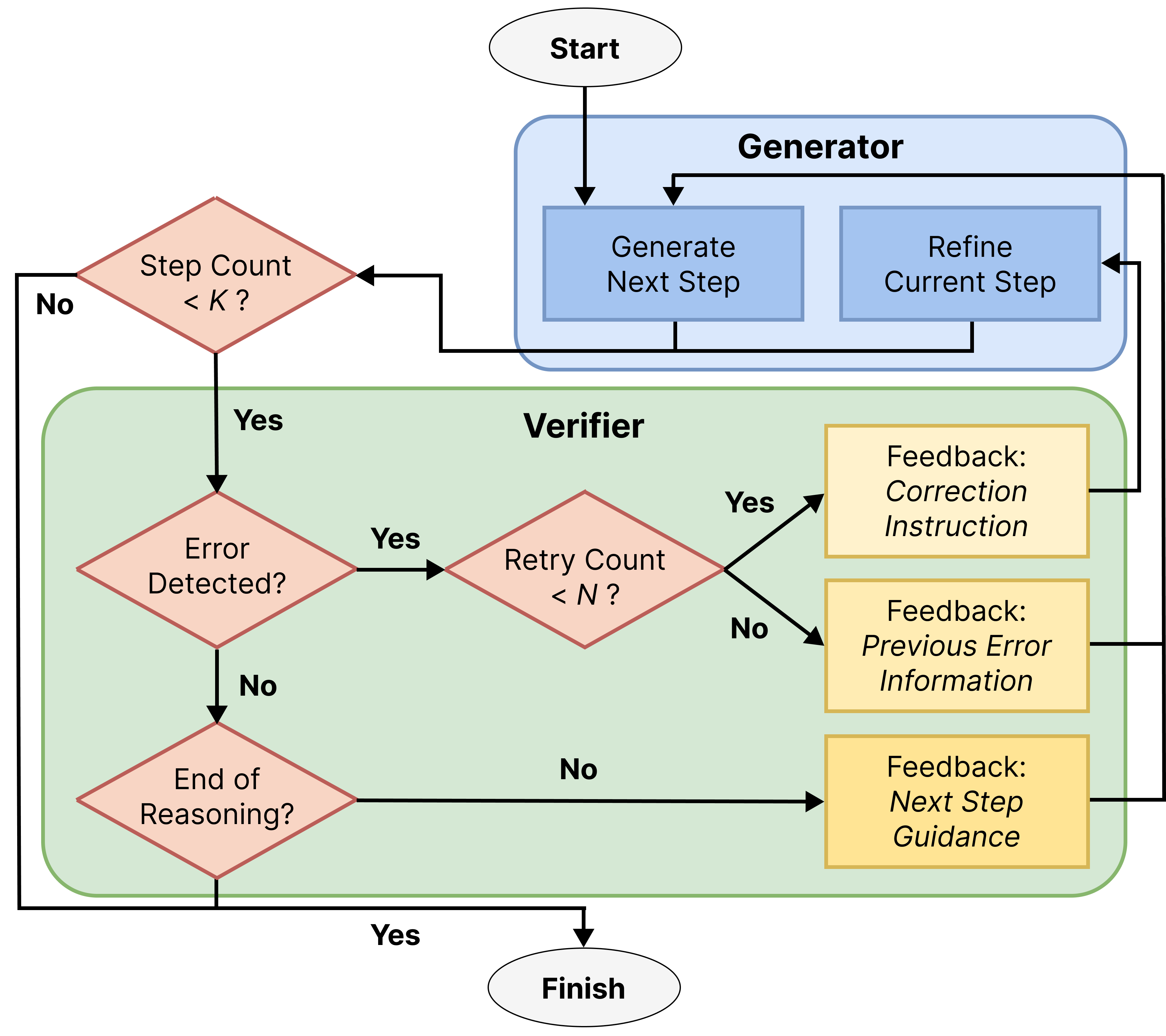}
    \vspace{-1.5em}
    \caption{Diagram of \ours{} at inference-time.
    The Generator produces or refines an atomic reasoning step, while the Verifier detects errors and provides feedback. This step-level feedback loop continues until the reasoning terminates.
}
    \label{fig:diagram}
}
\end{figure}

\subsection{Verifier Training Dataset}
\label{sec:appendix_training_dataset}

\begin{figure}[t!]
{
\centering
    \includegraphics[width=\linewidth]{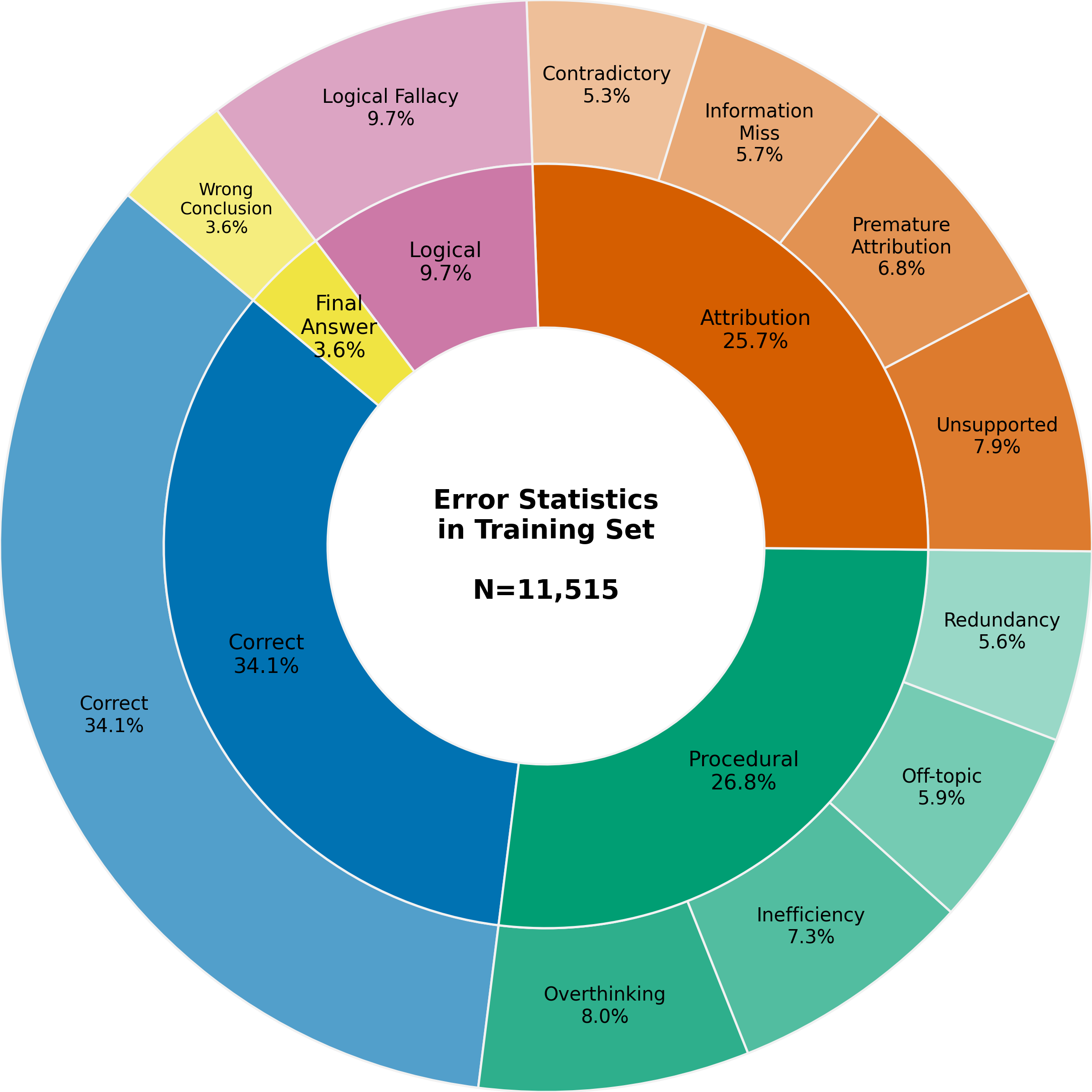}
    \caption{
Distribution of diagnostic error categories in the verifier training set. 
The inner ring shows broad categories, while the outer ring shows specific error types, illustrating coverage of diverse failure modes used for stepwise verification.
}
    \label{fig:training_dataset_statistics}
}
\end{figure}

This section describes the structure, composition, and statistical distribution of the dataset used to train our stepwise verifier.

Figure~\ref{fig:training_data_example} illustrates a representative training instance. 
Unlike post-hoc evaluation datasets, our training data is structured to support step-level verification during generation. 
Each instance provides the verifier with the question, the provided passages, the previous reasoning steps, and the current step to be evaluated. 
The target output follows the verifier response format used in \ours{}: an \textit{Error Type}, a natural language \textit{Diagnosis}, and actionable \textit{Guidance}. 
Together, these fields teach the verifier not only to detect unsupported reasoning steps, but also to explain the violation and guide the generator toward a valid correction.

Table~\ref{tab:training_data_unique_question} reports the statistics of the constructed training dataset across 2WikiMultihopQA, HotpotQA, and MuSiQue. 
The dataset contains 11.5k step-level verification instances derived from 5k unique questions. 
Because \ours{} verifies reasoning iteratively, a single question can yield multiple training instances corresponding to different reasoning prefixes and current steps. 
This one-to-many structure exposes the verifier to diverse valid and invalid steps within multi-hop reasoning trajectories.

\begin{figure}[t!]
{
\centering
    \includegraphics[width=\linewidth]{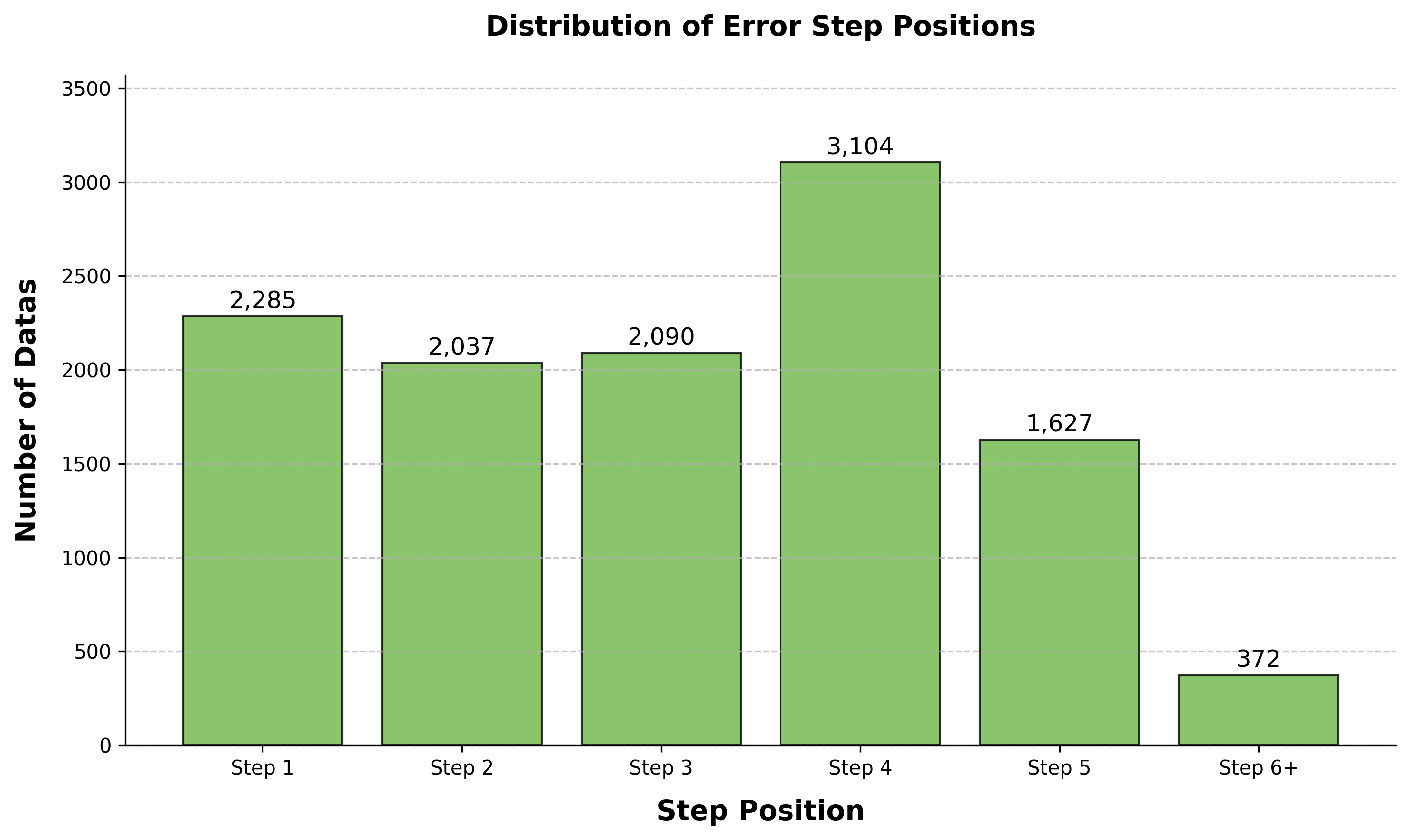}
    \caption{
Positional distribution of error steps in the verifier training data. 
Errors appear across different reasoning stages, providing supervision for detecting unsupported steps at multiple points during generation.
}
    \label{fig:error_position_statistics}
}
\end{figure}

Finally, Figure~\ref{fig:error_position_statistics} shows the positional distribution of error steps across reasoning trajectories. 
A stepwise verifier should be able to detect unsupported reasoning regardless of where it occurs in the trajectory, rather than only at the beginning or near the final answer. 
As shown in the figure, the training data covers errors across different reasoning positions. 
This distribution helps the verifier learn to diagnose and correct failures throughout the generation process.

\begin{figure}[t]
\centering
\begin{tcolorbox}[
    colback=gray!5,
    colframe=gray!75,
    title={Example Training Data},
    fonttitle=\bfseries,
    boxrule=0.5pt,
    arc=2pt,
    left=2mm,
    right=2mm,
    top=1mm,
    bottom=1mm,
    boxsep=2pt
]
\small

\textbf{Question:} Are director of film \textit{The Raft of the Dead} and director of film \textit{The World Gone Mad} both from the same country?

\par\noindent\rule{\linewidth}{0.4pt}\par

\textbf{Passages:}\\
\texttt{Passage 1:} The World Gone Mad ... is a 1933 American ... directed by Christy Cabanne ...\\
\texttt{Passage 5:} Carl Boese ... was a German film director ...\\
\texttt{Passage 8:} Christy Cabanne ... was an American film director ...\\
\texttt{Passage 9:} The Raft of the Dead ... directed by Carl Boese ...

\par\noindent\rule{\linewidth}{0.4pt}\par

\textbf{Previous Reasoning Steps:}\\
\texttt{Step 1:} Passage 9 states the director of \textit{The Raft of the Dead} is Carl Boese.\\
\texttt{Step 2:} Passage 5 reveals the country of origin of Carl Boese is Germany.\\
\texttt{Step 3:} Passage 1 reveals the director of \textit{The World Gone Mad} is Christy Cabanne.\\
\texttt{Step 4:} The country of origin of Christy Cabanne (from Step 3) is America, based on Passage 8.

\par\noindent\rule{\linewidth}{0.4pt}\par

\textbf{Current Step:}\\
\texttt{Step 5:} As previously established in Step 2, Carl Boese is indeed from Germany.

\par\noindent\rule{\linewidth}{0.4pt}\par

\textbf{Verifier Output:}\\
\texttt{Error Type:} Redundancy\\
\texttt{Diagnosis:} The step repeats the previously established fact from Step 2 without adding any new inference or moving closer to the final answer.\\
\texttt{Guidance:} Compare the countries of origin established in Step 2 (Germany) and Step 4 (America) to determine if they are the same.

\end{tcolorbox}

\vspace{-1em}
\caption{
Example verifier training instance. 
The verifier receives the question, provided passages, previous reasoning steps, and the current step, then outputs an error type, diagnosis, and corrective guidance. 
For brevity, distractor passages are omitted from the original 10-passage context, and step-type tags are removed.
}
\label{fig:training_data_example}
\end{figure}
\begin{table}[t]
\centering
\small
\begin{tabular}{lcc}
\toprule
\textbf{Dataset} & \textbf{Step-Level Instances} & \textbf{Unique Questions} \\
\midrule
2Wiki & 5,476 & 2,292 \\
HotpotQA & 3,063 & 1,330 \\
MuSiQue & 2,976 & 1,438 \\
\midrule
\textbf{Total} & \textbf{11,515} & \textbf{5,060} \\
\bottomrule
\end{tabular}
\caption{
Statistics of the verifier training dataset. 
Step-level instances correspond to individual verifier inputs constructed from reasoning prefixes and current steps. 
Because \ours{} verifies reasoning iteratively, a single question can produce multiple step-level training instances.
}
\label{tab:training_data_unique_question}
\end{table}

\subsection{Qualitative Case Studies}
\label{sec:appendix_case_study}

In this section, we provide qualitative case studies comparing \ours{} with the self-verification baseline. 
These examples illustrate how an external stepwise verifier can detect unsupported reasoning steps, provide diagnostic and actionable guidance, and help the generator remain grounded in the provided passages during generation.

Table~\ref{tab:qualitative_comparison} shows a case involving lexical distractors. 
The self-verification baseline is misled by lexical overlap with the term ``Eden'' in Passage 1. 
However, the relevant entity in the question is ``Lake Eden,'' which is supported by different evidence in Passage 10. 
\ours{} identifies this mismatch and guides the generator away from the incorrect entity. 
By distinguishing the hamlet ``Eden'' from the target entity ``Lake Eden,'' \ours{} keeps the reasoning process grounded in the correct evidence.

Table~\ref{tab:premature_attribution_appendix} illustrates a structural reasoning failure caused by premature attribution. 
In this example, the self-verification baseline jumps directly to extracting a death date without first establishing the required link between the film and the director. 
This early unsupported step leads to a chain of implicit assumptions and causes the model to lose track of the question's main constraint, which is to identify a \textit{film title}. 
As a result, the model outputs the director's name instead, even though the relevant passages are available.

In contrast, \ours{} detects the missing link at the first step and provides guidance for revising the reasoning path. 
By correcting this early unsupported step, \ours{} helps ensure that later steps are built on evidence-grounded reasoning. 
In the final stage, the verifier also provides guidance that the answer should be a \textit{film title}, keeping the reasoning process aligned with the question requirement.

Finally, Table~\ref{tab:self_feedback_collapse} shows a case where self-verification leads to a broader reasoning breakdown. 
The baseline exhibits target drift: instead of maintaining focus on the entity requested by the question, the second largest city Tucson, it shifts the generator toward an irrelevant target, Phoenix. 
After failing to find the population of Phoenix in the provided passages, the self-verifier further instructs the generator to rely on external sources. 
This example shows that self-verification can amplify errors when the evaluator itself is not grounded in the provided evidence. 
By separating generation from verification, \ours{} reduces this failure mode and keeps correction guidance tied to the available context.

\subsection{Ablation: Synthetic-Only vs. Refined Verifier Training}
\label{sec:appendix_synthetic_only_vs_full}

To evaluate the effect of targeted refinement in verifier training, we compare two variants: a verifier trained only on synthetic error-injected examples (``Synthetic Only'') and the final verifier trained with additional hard cases collected through our analysis-driven refinement strategy (``Refined''). 
Table~\ref{tab:synthetic_vs_refined} reports the results across generator models with different architectures and sizes.

Overall, adding refined training examples consistently improves performance, yielding an average gain of +2.9 pp across all evaluated models. 
The largest gains appear on MuSiQue, the most challenging benchmark in our evaluation, where the average improvement reaches +4.2 pp. 
This suggests that synthetic error injection provides broad coverage of diagnostic categories, while targeted refinement captures more complex reasoning failures that arise during actual inference.

At the model level, the improvement is most pronounced for Llama-3.1-8B, which gains +6.1 pp on average and +8.6 pp on MuSiQue. 
The refined data also improves Gemma-3-12B by +3.0 pp on average. 
The gains are smaller for Qwen2.5-14B, with a +0.4 pp average improvement, and Qwen3-4B shows a slight regression on HotpotQA of -0.2 pp. 
Despite these small variations, the overall trend shows that targeted refinement improves the verifier's ability to handle diverse inference-time failure modes.

This ablation shows that synthetic error injection alone is not sufficient for training the final verifier. 
Adding refinement examples helps the verifier generalize beyond controlled synthetic errors and better support step-level correction during generation.

\begin{table}[t]
\centering
\resizebox{\columnwidth}{!}{
\begin{tabular}{lcccc}
\toprule
\textbf{Component} & \textbf{Total} & \textbf{Cached} & \textbf{Computed} & \textbf{Savings} \\
\midrule
Generator & 43.8M & 32.9M & 10.9M & 75.2\% \\
Evaluator & 100.4M & 95.4M & 5.0M & 95.0\% \\
\midrule
\textbf{Overall} & \textbf{144.2M} & \textbf{128.3M} & \textbf{15.9M} & \textbf{89.0\%} \\
\bottomrule
\end{tabular}
}
\caption{
Inference cost measured in total and cached tokens. In \ours{}, prefix KV caching eliminates 89.0\% of redundant computations, and the marginal increase remains practically negligible.
}
\label{tab:kv_caching_efficiency}
\end{table}
\begin{table*}[h]
\centering
\small
\begin{tabularx}{\textwidth}{l X}
\toprule
\textbf{Question} & Are both Missisa Lake and Lake Eden located in the same country? \\ \midrule
\textbf{Passages} & [1] \textit{Eden is a CDP in Erie County, New York, United States.} (Distractor) \newline 
                        [6] \textit{Missisa Lake is a lake in Ontario, Canada.} \newline 
                        [10] \textit{Lake Eden is a small, recreational lake in Alberta, Canada.} \\ \midrule
\textbf{Reasoning Step} & \textbf{Step 1:} According to Passage 1, Lake Eden is located in New York. \\ \midrule \midrule
\textbf{Self-Verification} & \textbf{error\_type:} \textcolor{red}{Correct (No Error)} \newline 
                                   \textbf{diagnosis:} The step correctly extracts the location of Lake Eden from Passage 1. \newline 
                                   \textbf{guidance:} Find the location of Missisa Lake from the provided passages. \\ \midrule
\textbf{Ours} & \textbf{error\_type:} \textcolor{blue}{Contradictory} \newline 
                                   \textbf{diagnosis:} The step misreads Passage 10 by claiming Lake Eden is in New York, whereas Passage 10 explicitly states that Lake Eden is in Alberta, Canada. This directly contradicts the retrieved passage. \newline 
                                   \textbf{guidance:} Correct the attribution by stating that Lake Eden is located in Alberta, Canada, based on Passage 10. \\ \bottomrule
\end{tabularx}
\caption{
Qualitative comparison between self-verification and \ours{}. 
The self-verification baseline is misled by lexical overlap with the distractor term ``Eden'' in Passage 1 and fails to distinguish it from the target entity ``Lake Eden.'' 
In contrast, \ours{} detects the entity mismatch and provides evidence-grounded corrective guidance.
}
\label{tab:qualitative_comparison}
\end{table*}

\begin{table*}[h] 
\centering
\small
\begin{tabularx}{\textwidth}{X | X}
\toprule
\multicolumn{2}{p{\dimexpr\textwidth-2\tabcolsep\relax}}{
    \textbf{Question:} Which film has the director who died earlier, \textit{Un'Estate Ai Caraibi} or \textit{Her Honor, The Governor}? \vspace{0.5em} \newline
    \textbf{Passages (Truncated):} \newline
    [2] \textit{Chester ``Chet'' Withey... died 6 October 1939.} \newline
    [4] \textit{Her Honor, the Governor is a 1926... film starring Pauline Frederick, directed by Chester Withey...} \newline
    [5] \textit{Carlo Vanzina (13 March 1951 -- 8 July 2018) was an Italian film director, producer and screenwriter...} \newline
    [7] \textit{Un'estate ai Caraibi is a 2009 Italian comedy film directed by Carlo Vanzina...} 
} \cr \midrule \midrule

\multicolumn{1}{>{\centering\arraybackslash}X}{\textbf{Self-Verification}} & 
\multicolumn{1}{>{\centering\arraybackslash}X}{\textbf{Ours}} \cr \midrule

\textbf{Step 1:} According to Passage 2, Chester ``Chet'' Withey died on Oct 6, 1939. \vspace{0.3em} \newline
\textbf{error\_type:} \textcolor{red}{Correct (No Error)} \newline
\textbf{diagnosis:} The step correctly extracts the death date of Chester Withey. \newline
\textbf{guidance:} Find the death date of Carlo Vanzina. \vspace{0.8em} \newline
& 
\textbf{Step 1:} According to Passage 2, Chester ``Chet'' Withey died on Oct 6, 1939. \vspace{0.3em} \newline
\textbf{error\_type:} \textcolor{blue}{Premature Attribution} \newline
\textbf{diagnosis:} Missing the preceding step to identify that Withey is the director of ``Her Honor, The Governor''. \newline
\textbf{guidance:} First, explicitly identify the director of ``Her Honor, The Governor'' based on Passage 4. \cr \midrule

\textbf{Step 2:} According to Passage 5, Carlo Vanzina died on July 8, 2018. (Accepted) \vspace{0.5em} \newline
\textbf{Step 3:} Comparing the death date of Chester Withey (October 6, 1939) with the death date of Carlo Vanzina (July 8, 2018), Chester Withey died earlier. (Accepted)
& 
\textbf{Step 1 (Retry):} \newline
According to Passage 4, Chester Withey directed ``Her Honor, The Governor''. (Accepted) \vspace{0.5em} \newline
\textbf{Step 2-4:} Identifies Vanzina as director, then collects death dates for both directors. (Accepted) \cr \midrule

\textbf{Step 4:} \textcolor{red}{\#\#\#\#ANSWER: Chester Withey} \vspace{0.5em} \newline
\textbf{Analysis:} Because the feedback model did not enforce the director-to-film connection in Step 1, the generator lost track of the final goal (finding the \textit{film title}) and mistakenly output the \textit{director's name} instead.
& 
\textbf{Step 5:} Compares death dates of two directors. \vspace{0.5em} \newline
\textbf{Step 6:} \textcolor{blue}{\#\#\#\#ANSWER: Her Honor, The Governor} \vspace{0.5em} \newline
\textbf{Analysis:} The verifier's guidance explicitly instructed: \textit{``State which film has the director who died earlier... using format \#\#\#\#ANSWER: <film\_title>.''}. This ensured task compliance and correct entity type. \cr \bottomrule
\end{tabularx}
\caption{
Detailed trace comparison between self-verification and \ours{}. 
Self-verification loses entity consistency through premature attribution, whereas \ours{} detects the missing link and provides stepwise corrective guidance to keep the reasoning path evidence-grounded and aligned with the required answer type.
}
\label{tab:premature_attribution_appendix}
\end{table*}

\begin{table*}[h]
\centering
\small
\begin{tabularx}{\textwidth}{X}
\toprule
\multicolumn{1}{p{\dimexpr\textwidth-2\tabcolsep\relax}}{
    \textbf{Question:} In 1900, what was the population of the second largest city in the state where the Oh Yeah performer is from? \vspace{0.3em} \newline
    \textbf{Passages (Truncated):} \newline
    [2] \textit{Tucson is the largest city in southern Arizona, the second largest in the state after Phoenix.} \newline
    [6] \textit{By 1900, 7,531 people lived in the city Tucson.} \newline
    [9] \textit{Oh Yeah is a 1962 album by jazz musician Charles Mingus.} \newline
    [10] \textit{Charles Mingus was born in Nogales, Arizona.} \vspace{0.3em} \newline
    \textbf{Ground Truth Answer:} 7531
} \cr \midrule \midrule

\textbf{Self-Verification Reasoning Trace (Summarized)} \cr \midrule

\textbf{Step 4:} According to Passage 2, Phoenix is the second largest city in Arizona after Tucson. \\[0.3em]
\textbf{error\_type:} \textcolor{red}{Off-topic} \newline
\textbf{diagnosis:} The question asks for the population of the *second largest* city... While this step correctly identifies \textcolor{red}{Phoenix as the second largest city in Arizona}, it doesn't address the core requirement of finding the population in 1900. \newline
\textbf{guidance:} \textcolor{red}{Find the population of Phoenix} in 1900 from the passages. \cr \midrule

\textbf{Step 9:} According to Passage 2, Phoenix is the largest city in Arizona, but it does not state the population of Phoenix in 1900. \\[0.3em]
\textbf{error\_type:} Redundancy \newline
\textbf{diagnosis:} This step simply reiterates information already established in Step 8. \newline
\textbf{guidance:} \textcolor{red}{Find the population of Phoenix in 1900 from external sources.} \cr \midrule

\textbf{Step 10:} According to the \textcolor{red}{United States Census Bureau}, the population of Phoenix, Arizona in 1900 was \textcolor{red}{10611}. \cr \midrule

\textbf{Predicted Answer:} \textcolor{red}{10611} \cr \bottomrule
\end{tabularx}
\caption{
Detailed trace of reasoning collapse in the self-verification baseline. 
The self-verifier shifts the target from the requested entity, \textit{Tucson}, to an irrelevant entity, \textit{Phoenix}, and then instructs the generator to use external knowledge when the required evidence is not found in the provided passages. 
This example illustrates how self-verification can amplify unsupported reasoning when the evaluator is not grounded in the evidence.
}
\label{tab:self_feedback_collapse}
\end{table*}
\begin{table*}[t]
\centering
\small
\resizebox{0.95\linewidth}{!}{%
\begin{tabular}{llcccc}
\toprule
\textbf{Model} & \textbf{Verifier Training} & \textbf{2Wiki} & \textbf{HotpotQA} & \textbf{MuSiQue} & \textbf{Average} \\
\midrule

\multirow{2}{*}{Gemma-3-12B} 
 & Synthetic Only & 88.6 & 89.4 & 71.6 & 83.2 \\
 & + Refined & \textbf{91.8} \textcolor[HTML]{009B00}{\footnotesize (+3.2)} & \textbf{90.0} \textcolor[HTML]{009B00}{\footnotesize (+0.6)} & \textbf{76.7} \textcolor[HTML]{009B00}{\footnotesize (+5.1)} & \textbf{86.2} \textcolor[HTML]{009B00}{\footnotesize (+3.0)} \\
\midrule

\multirow{2}{*}{Qwen2.5-14B} 
 & Synthetic Only & \textbf{90.2} & 89.7 & 75.7 & 85.2 \\
 & + Refined & \textbf{90.2} \textcolor[HTML]{009B00}{\footnotesize (+0.0)} & \textbf{90.1} \textcolor[HTML]{009B00}{\footnotesize (+0.4)} & \textbf{76.6} \textcolor[HTML]{009B00}{\footnotesize (+0.9)} & \textbf{85.6} \textcolor[HTML]{009B00}{\footnotesize (+0.4)} \\
\midrule

\multirow{2}{*}{Llama-3.1-8B} 
 & Synthetic Only & 86.5 & 84.8 & 66.7 & 79.3 \\
 & + Refined & \textbf{90.7} \textcolor[HTML]{009B00}{\footnotesize (+4.2)} & \textbf{90.3} \textcolor[HTML]{009B00}{\footnotesize (+5.5)} & \textbf{75.3} \textcolor[HTML]{009B00}{\footnotesize (+8.6)} & \textbf{85.4} \textcolor[HTML]{009B00}{\footnotesize (+6.1)} \\
\midrule

\multirow{2}{*}{Qwen3-8B} 
 & Synthetic Only & 88.6 & 88.6 & 70.9 & 82.7 \\
 & + Refined & \textbf{91.3} \textcolor[HTML]{009B00}{\footnotesize (+2.7)} & \textbf{90.9} \textcolor[HTML]{009B00}{\footnotesize (+2.3)} & \textbf{76.6} \textcolor[HTML]{009B00}{\footnotesize (+5.7)} & \textbf{86.3} \textcolor[HTML]{009B00}{\footnotesize (+3.6)} \\
\midrule

\multirow{2}{*}{Qwen3-4B} 
 & Synthetic Only & 87.3 & \textbf{89.2} & 73.5 & 83.3 \\
 & + Refined & \textbf{89.4} \textcolor[HTML]{009B00}{\footnotesize (+2.1)} & 89.0 {\scriptsize (-0.2)} & \textbf{74.5} \textcolor[HTML]{009B00}{\footnotesize (+1.0)} & \textbf{84.3} \textcolor[HTML]{009B00}{\footnotesize (+1.0)} \\
\midrule
\midrule

\multirow{2}{*}{\textbf{Average}} 
 & Synthetic Only & 88.2 & 88.3 & 71.7 & 82.7 \\
 & + Refined & \textbf{90.7} \textcolor[HTML]{009B00}{\footnotesize (+2.5)} & \textbf{90.1} \textcolor[HTML]{009B00}{\footnotesize (+1.8)} & \textbf{75.9} \textcolor[HTML]{009B00}{\footnotesize (+4.2)} & \textbf{85.6} \textcolor[HTML]{009B00}{\footnotesize (+2.9)} \\
\bottomrule
\end{tabular}%
}
\caption{
Ablation of targeted refinement for verifier training. 
``Synthetic Only'' uses only diagnostic error-injected examples, while ``+ Refined'' adds hard cases collected through analysis-driven targeted refinement. 
Values in parentheses indicate absolute changes relative to Synthetic Only.
}
\label{tab:synthetic_vs_refined}
\end{table*}
\begin{table*}[t]
\centering
\small
\resizebox{0.8\textwidth}{!}{%
\begin{tabular}{llcccc}
\toprule
\textbf{Model} & \textbf{Method} & \textbf{2Wiki} & \textbf{HotpotQA} & \textbf{MuSiQue} & \textbf{Average} \\
\midrule

\multirow{3}{*}{Gemma 3 12B}
& No Verification   & 76.7 & 82.4 & 62.8 & 74.0 \\
& Self Verification & 71.2 & 80.3 & 64.6 & 72.0 \\
& \textbf{SAFE (Ours)}     & \textbf{91.2} & \textbf{90.0} & \textbf{76.1} & \textbf{85.8} \\
\midrule

\multirow{3}{*}{Qwen 2.5 14B}
& No Verification   & 83.3 & 84.6 & 63.0 & 77.0 \\
& Self Verification & 78.7 & 82.8 & 57.9 & 73.1 \\
& \textbf{SAFE (Ours)}     & \textbf{89.4} & \textbf{88.1} & \textbf{75.2} & \textbf{84.2} \\
\midrule

\multirow{3}{*}{Llama 3 8B}
& No Verification   & 65.1 & 75.5 & 50.7 & 63.8 \\
& Self Verification & 58.8 & 68.0 & 41.8 & 56.2 \\
& \textbf{SAFE (Ours)}     & \textbf{89.8} & \textbf{90.0} & \textbf{74.1} & \textbf{84.6} \\
\midrule

\multirow{3}{*}{Qwen 3 8B}
& No Verification   & 82.9 & 83.3 & 63.0 & 76.4 \\
& Self Verification & 80.2 & 81.6 & 62.0 & 74.6 \\
& \textbf{SAFE (Ours)}     & \textbf{90.0} & \textbf{90.5} & \textbf{75.8} & \textbf{85.4} \\
\midrule

\multirow{3}{*}{Qwen 3 4B}
& No Verification   & 81.7 & 83.5 & 59.7 & 75.0 \\
& Self Verification & 80.9 & 80.6 & 55.2 & 72.2 \\
& \textbf{SAFE (Ours)}     & \textbf{89.8} & \textbf{89.6} & \textbf{75.0} & \textbf{84.8} \\
\bottomrule
\end{tabular}%
}
\caption{LLM-judge accuracy with Llama-3.3-70B-Instruct.}
\label{tab:llama_judge}
\end{table*}

\subsection{Efficiency and Failure Patterns}
\label{sec:appendix_cost}

\paragraph{SAFE vs. Self-Verification.}

\begin{figure}[t!]
{
\centering
    \includegraphics[width=\linewidth]{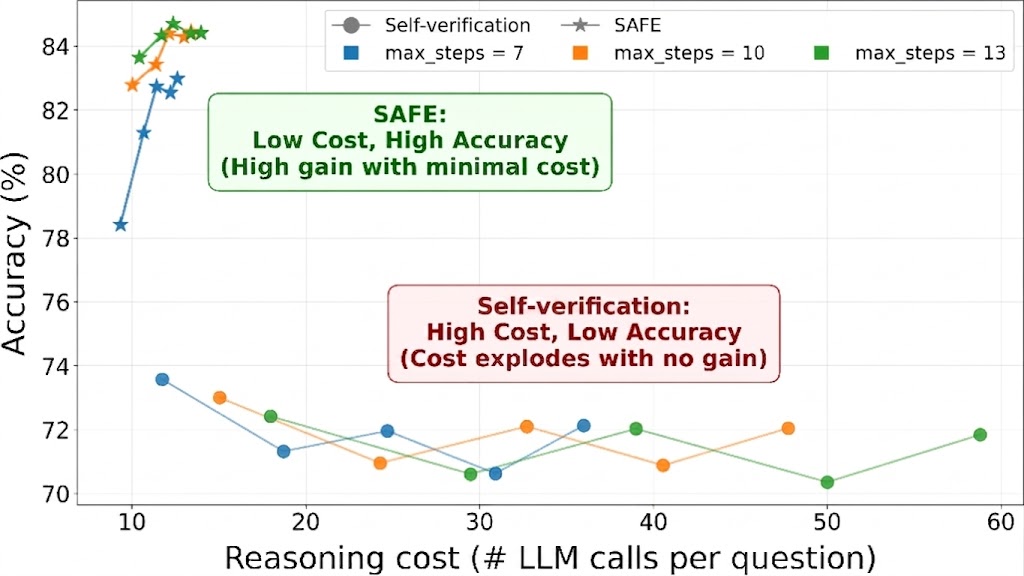}
    \vspace{-1em}
    \caption{
    Cost-accuracy comparison between \ours{} and the \textit{Self-Verification} baseline. 
    We evaluate the inference efficiency across different max\_steps ($K\in\{7, 10, 13\}$) and max\_retries ($N\in\{1..5\}$). While the \textit{Self-Verification} exhibits a severe cost explosion without meaningful accuracy improvements, \ours{} achieves superior accuracy with minimal computational overhead.
    }
    \label{fig:self_feedback_vs_safe}
}
\end{figure}

Figure~\ref{fig:self_feedback_vs_safe} compares the cost-accuracy trade-off between \ours{} and the self-verification baseline under different retry budgets. 
As the retry limit increases, self-verification requires substantially more LLM calls but does not achieve comparable accuracy gains. 
This occurs because the generator often fails to identify its own unsupported step and may repeatedly revise along an invalid trajectory. 
In contrast, \ours{} reaches over 84\% average accuracy with fewer than 15 calls on average. 
This suggests that external stepwise verification is more efficient than unguided self-correction for evidence-grounded multi-hop QA.

\paragraph{Cost-Accuracy Trade-off of \ours{}.}

\begin{figure}[t!]
{
\centering
    \includegraphics[width=\linewidth]{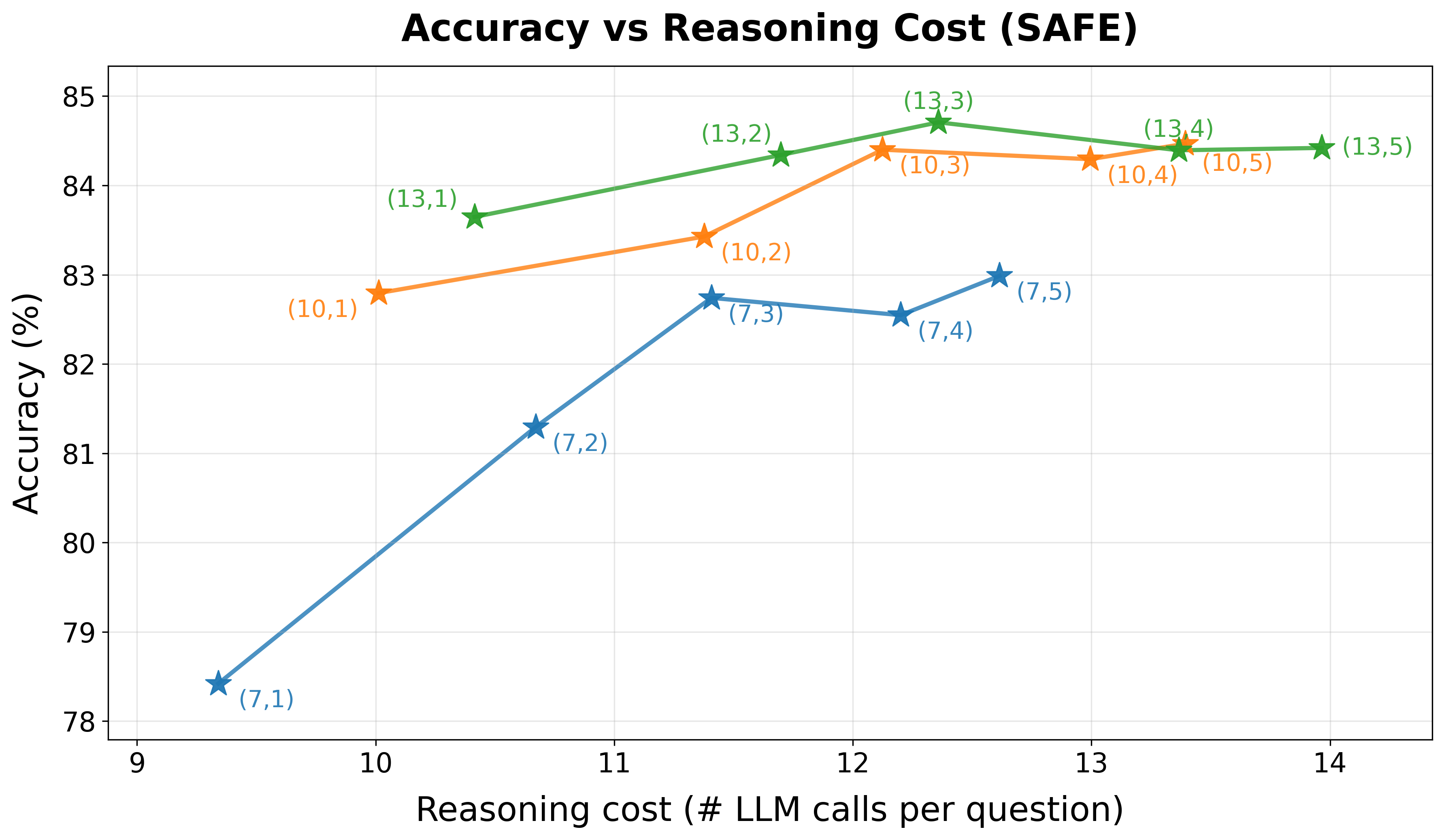}
    \vspace{-1em}
    \caption{Cost-accuracy trade-off of \ours{}. Annotations indicate the $(K, N)$ configuration, where $K$ is max\_steps and $N$ is max\_retries.}
    \label{fig:safe_cost_accuracy_analysis}
}
\end{figure}

To further analyze the cost-performance trade-off of inference-time verification, we vary the maximum number of reasoning steps $K$ and the maximum number of retry attempts $N$. 
As shown in Figure~\ref{fig:safe_cost_accuracy_analysis}, accuracy generally improves when the generator is given more steps and retry opportunities, since the verifier has more chances to detect unsupported steps and guide corrections. 
However, performance begins to plateau as the retry budget increases. 
Increasing $N$ beyond 3 yields only marginal gains and can sometimes slightly degrade performance.

The highest accuracy, 84.71\%, is achieved with $K=13$ and $N=3$, but this setting requires additional computation. 
We therefore use $K=10$ and $N=3$ as the default configuration, which achieves a similar accuracy of 84.40\% with a lower reasoning budget. 
This setting provides a practical balance between verification cost and answer accuracy.

\paragraph{Prefix KV Caching.}

A key concern with iterative inference-time verification is the additional computational cost introduced by repeated generation and verification. 
In multi-hop QA, much of this cost comes from the prefill stage, where the model processes the system instruction, provided passages, question, and previous reasoning context.

As shown in Table~\ref{tab:kv_caching_efficiency}, \ours{} mitigates this overhead through Prefix KV caching~\citep{kwon2023efficient}. 
Both the reasoning generator and the stepwise verifier independently cache the representations of static prefixes, such as the instruction, passages, and question. 
As generation proceeds, verified reasoning steps are appended to the cached context. 
Thus, later reasoning and verification calls mainly require computation for newly generated tokens rather than reprocessing the entire prefix.

Across 3,000 evaluation examples from the three benchmarks, Prefix KV caching reduces redundant prefill token processing by 89.0\%. 
This result shows that \ours{} can support iterative stepwise verification with substantially reduced computational overhead.

\paragraph{Failure Type Distribution.}

\begin{figure}[t!]
\centering
    \includegraphics[width=\linewidth]{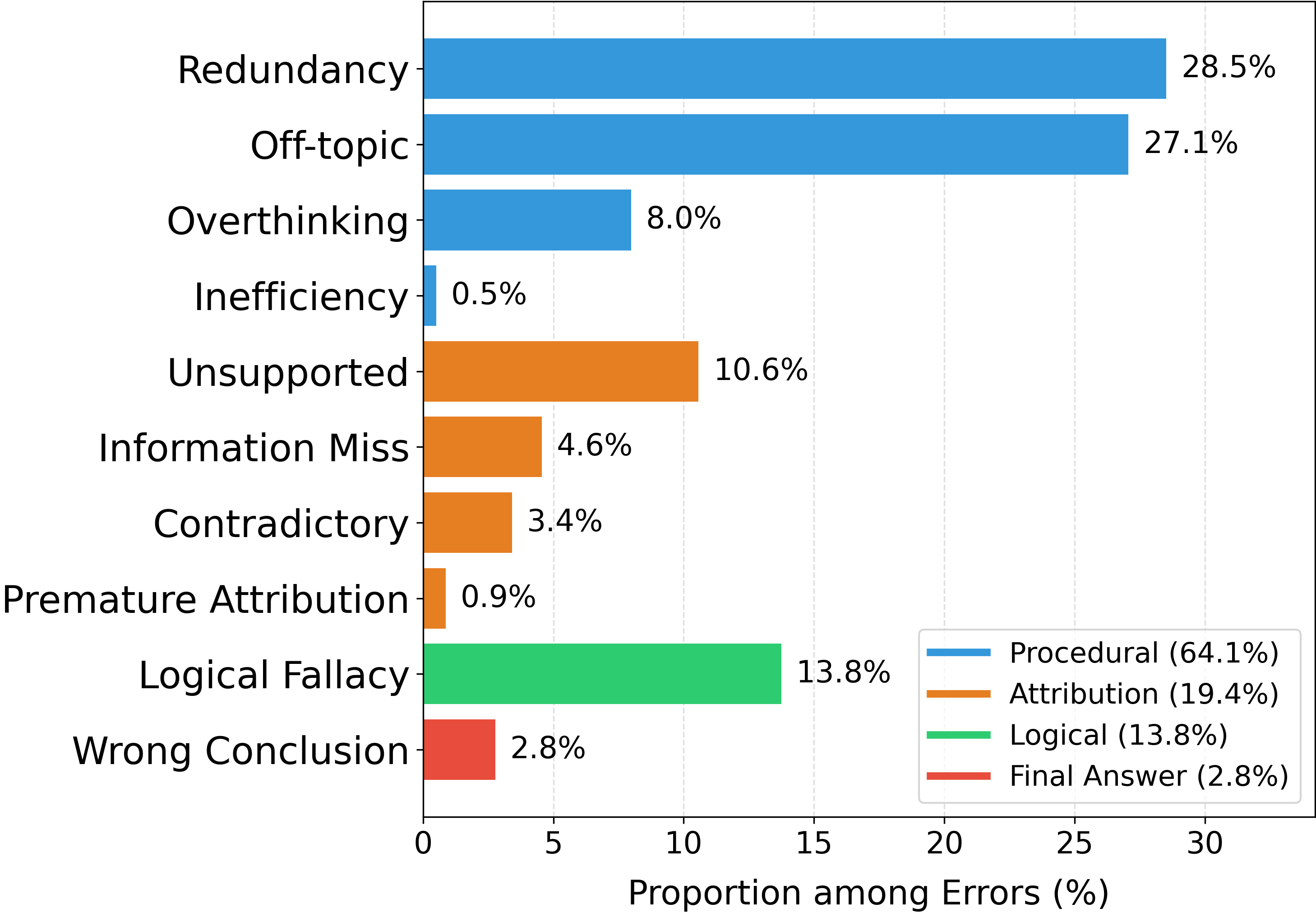}
    \vspace{-1.5em}
    \caption{
    Distribution of error types detected by \ours{}.
    Procedural errors, such as Redundancy and Off-topic, account for the majority of failures (64.1\%), suggesting that many errors arise from invalid reasoning structure rather than missing factual knowledge.
    }
    \label{fig:reasoning_error_distribution}
\end{figure}

We also analyze the failures detected by the verifier. 
Figure~\ref{fig:reasoning_error_distribution} shows that Procedural errors, such as redundancy and off-topic reasoning, account for 64.1\% of detected failures. 
This indicates that many multi-hop reasoning failures are not simply caused by missing factual knowledge, but by invalid reasoning structure. 
This finding further supports the need for a verifier that checks not only factual attribution, but also the structure and progress of the reasoning path.

\paragraph{Error Position Distribution.}

\begin{figure}[t!]
\centering
    \includegraphics[width=\linewidth]{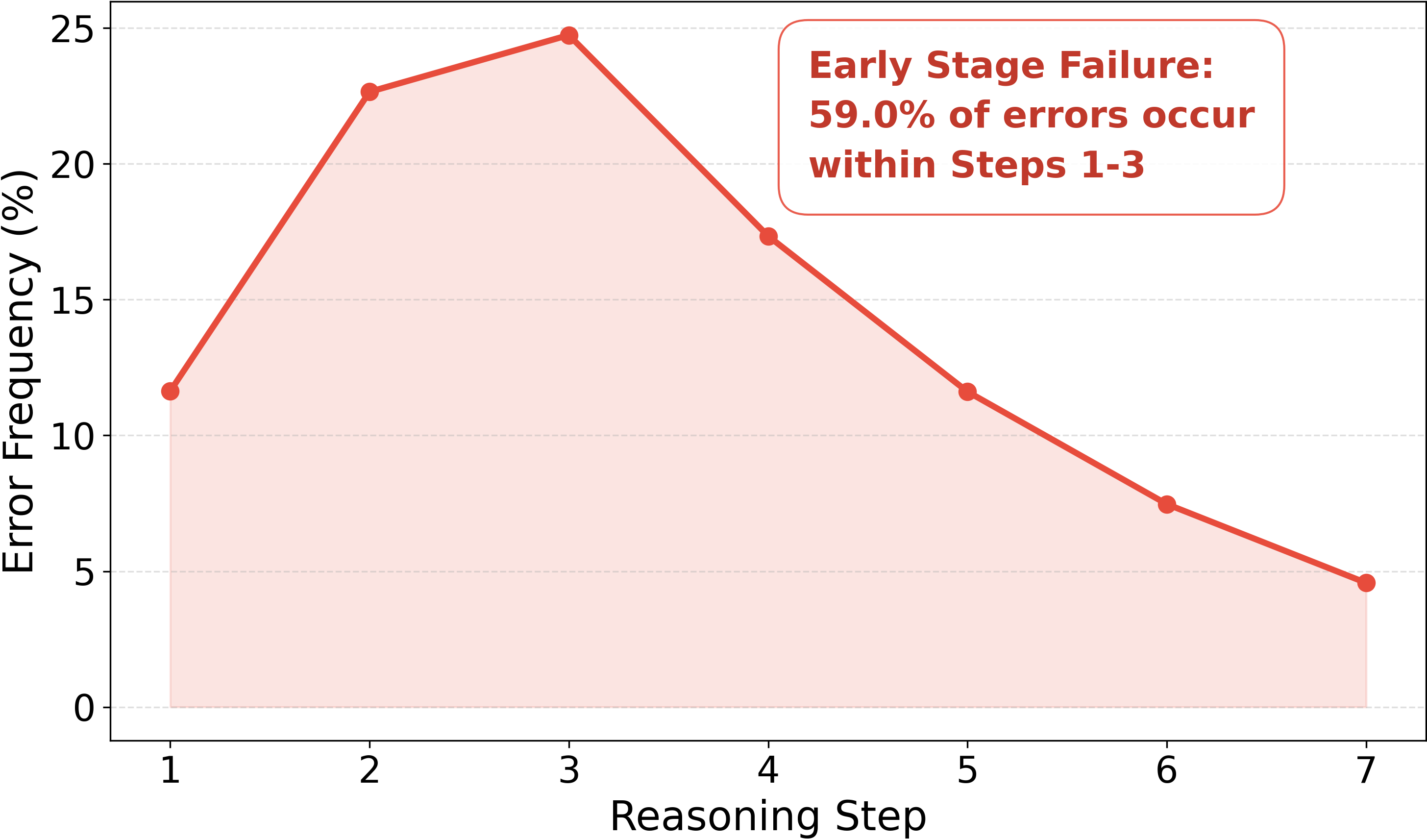}
    \vspace{-1.5em}
    \caption{
    Distribution of detected errors across reasoning steps.
    A large fraction of failures occurs within the first three steps (59.0\%), supporting the use of inference-time verification to detect and correct unsupported steps before they propagate.
    }
    \label{fig:error_step_position_safe}
\end{figure}

Figure~\ref{fig:error_step_position_safe} shows that 59.0\% of detected errors occur within the first three reasoning steps. 
This early concentration helps explain why inference-time verification is useful. 
Unsupported early steps can shape the rest of the reasoning trajectory, so correcting them before generation continues helps reduce downstream error propagation.

\paragraph{Qualitative Failure Modes.}

Qualitative examples in Appendix~\ref{sec:appendix_case_study} further illustrate the limitations of self-verification. 
Self-verification often fails due to lexical overlap bias, unsupported logical jumps, and evaluator hallucination. 
In contrast, \ours{} separates generation from verification and uses evidence-grounded diagnosis with actionable guidance, allowing the generator to revise unsupported steps before continuing.

\subsection{Prompts}
\label{sec:appendix_prompts}

This section provides the prompt templates used throughout \ours{} to support transparency and reproducibility. 
For clarity, we group the prompts according to their roles in the framework:

\begin{itemize}[leftmargin=*]
    \item \textbf{Inference-Time Verification and Answer Evaluation:} 
    This group includes prompts for step-by-step reasoning generation (Figure~\ref{fig:reasoning_step_generation_prompt}), inference-time step verification and corrective guidance (Figure~\ref{fig:evaluation_prompt}), and final answer extraction (Figure~\ref{fig:final_answer_generation_prompt}). 
    It also includes the LLM-as-judge prompt used to evaluate semantic equivalence between the predicted answer and the ground-truth answer (Figure~\ref{fig:llm_judge_prompt}).

    \item \textbf{Verifier Training Data Construction:} 
    This group contains prompts for dataset-specific query decomposition and reasoning planning for 2WikiMultihopQA (Figure~\ref{fig:plan_generation_prompt_2wiki}), HotpotQA (Figure~\ref{fig:plan_generation_prompt_hotpotqa}), and MuSiQue (Figure~\ref{fig:plan_generation_prompt_musique}). 
    It also includes prompts for generating evidence-grounded reasoning trajectories (Figure~\ref{fig:ideal_reasoning_generation_prompt}) and creating negative examples through diagnostic error injection (Figure~\ref{fig:error_injection_prompt}).

    \item \textbf{KG-Grounded Benchmark Verification:} 
    This group details the prompts used in our KG-grounded verification pipeline to identify unverifiable benchmark supervision. 
    The prompts cover triple extraction (Figures~\ref{fig:triple_extraction_prompt} and~\ref{fig:triple_gleaning_prompt}), entity resolution for synonymous or coreferential mentions (Figure~\ref{fig:entity_resolution_prompt}), and reasoning path discovery for checking whether the question entities can be connected to the answer through evidence-supported triples (Figure~\ref{fig:logical_path_discovery_prompt}).
\end{itemize}

\begin{figure*}[t]
\begin{tcolorbox}[
    colback=gray!5,
    colframe=black,
    title=Reasoning Step Generation Prompt,
    fonttitle=\bfseries,
    fontupper=\small,
    sharp corners,
    boxrule=0.5pt
]
You are a meticulous, step-by-step logical reasoner. Your task is to solve a complex question by generating **ONLY THE NEXT SINGLE, ATOMIC STEP** in a chain of thought.
\\\\
\#\# Core Task Definition \\
You must analyze the `Question`, `Retrieved Passages`, and `Previous Reasoning Steps`.
Most importantly, you must analyze the `Feedback` received on the last step, which consists of three parts:\\
1. **Error Type**: The category of the error.\\
2. **Diagnosis**: An explanation about the latest previous step.\\
3. **Guidance**: Specific instruction on what to do in this current step. (e.g. Fixing previous step's error, Proceeding to extract new information, or Making the final logical conclusion).
\\
You must prioritize the **Guidance**. It tells you exactly what action to take now.

---

\#\# Step Classifications\\
Every step must be strictly classified into one of three types.
Attribution and Logical actions cannot be mixed in a single step.\\

\#\#\# 1. Attribution Step\\
- **Definition**: Extracts **ONE** explicit fact from a **SINGLE** retrieved passage.\\
- **Requirement**: You MUST explicitly cite the source (e.g., "According to Passage X...").\\
- **Constraint**: Do **NOT** combine information from multiple passages (e.g., "Passage 1 says X and Passage 2 says Y").\\
- **Format suffix**: End the sentence with `(Attribution)`.\\

\#\#\# 2. Logical Step
- **Definition**: Performs **ONE** logical operation (comparison, calculation, or inference) based **ONLY** on `Previous Reasoning Steps`.\\
- **Requirement**: Do NOT look up new information from passages.\\
- **Constraint**: This step is for **Intermediate Reasoning** only.\\
    - You must NOT output the final answer marker here.\\
    - Even if you derived the answer mentally, just state the conclusion of the logic (e.g., "A is older than B").\\
- **Format suffix**: End the sentence with `(Logical)`.\\

\#\#\# 3. Final Answer Step\\
- **Definition**: Submits the final answer.\\
- **Strict Syntax Rule**: This step must ONLY generate the final answer following the format: `\#\#\#\#ANSWER: <answer\_value>`.\\
- **Constraint**: Do not write "Therefore...", or "The answer is...". Just the marker and value.\\
- **Format suffix**: End with `(Final Answer)`.

---

\#\# Strict Formatting Rules\\
1. **Numbering**: Start your response with `Step K:`, where `K` is the next integer after the last step number.\\
2. **Atomic Nature**: Adhere strictly to the "One Step = One Action" rule defined above.\\
3. **Suffix Mandatory**: Every step must end with one of `(Attribution)`, `(Logical)`, `(Final Answer)`.\\
4. **Final Answer Pattern**: The pattern for the final step is immutable: `Step K: \#\#\#\#ANSWER: <answer\_value> (Final Answer)`

---

\#\# Examples of Valid Atomic Steps

[EXAMPLES...]\\
  
**FOR THE FINAL ANSWER STEP, YOU MUST USE EXACT FOLLOWING FORMAT:**\\
Step X: \#\#\#\#ANSWER: your\_answer\_here (Final Answer)
\end{tcolorbox}
\caption{Reasoning step generation prompt.}
\label{fig:reasoning_step_generation_prompt}
\end{figure*}

\begin{figure*}[t]
\begin{tcolorbox}[
    colback=gray!5,
    colframe=black,
    title=Reasoning Step Evaluation Prompt 1,
    fonttitle=\bfseries,
    fontupper=\small,
    sharp corners,
    boxrule=0.5pt
]
\# Role\\
You are a Precision Reasoning Evaluator. Your goal is to critically assess a specific reasoning step (Step to evaluate) within the context of a multi-hop QA task. You must verify if the step is logically sound, factually grounded in the provided `Retrieved Passages`, and efficiently moves towards the answer.\\

\# Input Data Context\\
- **Question**: The main query to answer.\\
- **Retrieved Passages**: The only source of truth. External knowledge is strictly forbidden.\\
- **Previous Steps**: The chain of thought leading up to the Step to evaluate.\\
- **Step to evaluate**: The specific step you need to evaluate. It must end with one of these tags:\\
  - `(Attribution)`: Extracting facts directly from a passage.\\
  - `(Logical)`: Intermediate reasoning (comparing, calculating) without the final answer marker.\\
  - `(Final Answer)`: Strict submission of the final answer ONLY (Format: `\#\#\#\#ANSWER: <answer\_value>`).\\

\# Task\\
1. Compare the `Step to evaluate` against the `Retrieved Passages` and `Previous Steps`.\\
2. Choose one `error\_type` from the given categories based on the Evaluation Protocol.\\
3. Generate a structured evaluation with `diagnosis` and `guidance`.\\

\# Feedback Guidelines

You must follow this Evaluation Protocol sequentially to determine the `error\_type`.\\

\#\# Phase 1: Assess (Final Answer) Steps\\
**Condition**: If the `Step to evaluate` is tagged as `(Final Answer)`.\\
- **Check 1 (Consistency)**: Does the submitted answer value match the conclusion derived from the preceding steps?\\
    - If NO -> error\_type: Wrong Conclusion\\
- **Check 2 (Correctness)**: Are sufficiency and consistency met?\\
    - If YES -> error\_type: Correct (No Error)\\

\#\# Phase 2: Assess Utility \& Progress (For Attribution/Logical Steps)\\
**Condition**: If the `Step to evaluate` is `(Attribution)` or `(Logical)`.\\
- **Check 1 (Necessity)**: Can the final answer be fully derived only from previous steps?\\
    - If YES -> error\_type: Overthinking\\
- **Check 2 (Relevance)**: Is this step deals with necessary information to answer the question?\\
    - If NO (e.g., deriving true but useless facts, focusing on wrong entities) -> error\_type: Off-topic\\
- **Check 3 (Novelty)**: Does this step provide new meaningful information or deduction not present in previous steps?\\
    - If NO -> error\_type: Redundancy\\
- **Check 4 (Efficiency)**: Does this step actually perform a meaningful action (extraction/deduction)?\\
    - If NO (e.g., purely planning, stating "I will now...", or summarizing without progress) -> error\_type: Inefficiency\\

\#\# Phase 3: Assess Validity \& Soundness (For Attribution/Logical Steps)\\
**Condition**: If the step passes Phase 2 (it is useful and relevant), now check its truthfulness.\\
**[If Attribution Step]**\\
- **Check 1 (Consistency)**: Does it contradict the Passage?\\
    - If YES -> error\_type: Contradictory\\
- **Check 2 (Grounding)**: Is the fact explicitly present in the referenced Passage?\\
    - If NO (Hallucination) -> error\_type: Unsupported\\
- **Check 3 (Completeness)**: Does it claim information is missing when the Passage actually has it?\\
    - If YES -> error\_type: Information Miss\\
- **Check 4 (Ordering)**: Does this step extract an attribute (e.g., nationality, birth date) of an entity before establishing the necessary relationship (e.g., "is the director of...") that connects this entity to the question's subject?\\
    - If YES -> error\_type: Premature Attribution\\

**[If Logical Step]**\\
- **Check 1 (Soundness)**: Is the calculation, comparison, or inference logically valid?\\
    - If NO -> error\_type: Logical Fallacy\\

\#\# Priority Rules\\
This protocol is hierarchical. You must stop at the first error type with highest priority.\\
1. **Phase 1 (Final Answer Checks)** take precedence over everything else for `(Final Answer)` steps.\\
2. **Phase 2 (Utility Checks)** take precedence over Phase 3.\\
   - If a step is useless (e.g., Redundant, Off-topic, Overthinking, Inefficiency), it is an error regardless of whether it is factually true or false.\\
   - Do NOT check for Hallucinations (Phase 3) if the step has already failed a Utility Check (Phase 2).\\
   - Report ONLY the first error encountered.
\end{tcolorbox}
\end{figure*}

\begin{figure*}[t]
\begin{tcolorbox}[
    colback=gray!5,
    colframe=black,
    title=Reasoning Step Evaluation Prompt 2,
    fonttitle=\bfseries,
    fontupper=\small,
    sharp corners,
    boxrule=0.5pt
]
\# Output Generation Instructions\\

After determining the `error\_type` using the Evaluation Protocol (Phase 1-3), you must generate the `diagnosis` and `guidance` fields following these rules.\\

\#\# 1. How to Write "Diagnosis"\\
The diagnosis must be a self-contained explanation of *why* the specific `error\_type` was chosen.\\
NO Protocol References: DO NOT explicitly mention "Phase 1", "Phase 2", "Check 1", etc. The protocol is for your internal reasoning only. In the output, describe the content issue directly.\\
Be concise and avoid verbosity. Get straight to the point. Do not repeat the entire content of the step.\\

- **If Error**:\\
    - **Cite the Violation**: Explicitly mention which Check in the Protocol failed.\\
    - **Provide Evidence**: Quote conflicting text, state missing facts, or compare derived vs. submitted values.\\
- **If Correct**:\\
    - Briefly explain the specific contribution of this step to the overall reasoning chain.\\

\#\# 2. How to Write "Guidance"\\
Based on your `diagnosis`, provide a concise, specific instruction for the **single next immediate step**:\\
- **If the Step to evaluate has an Error**: Explicitly instruct how to fix the error in the immediate next step.\\
- **If the Step to evaluate is Correct**: Instruct the specific reasoning action required for the next step.\\

Important: The guidance must focus ONLY on the single, atomic next action. Do not provide a long-term plan or list multiple future steps (e.g., "Do A, then B, then C"). Just tell the model to do "A".\\

If your guidance instruct to generate the final answer step, your guidance must say to include the exact format required: `\#\#\#\#ANSWER: <answer\_value>`.

---

\# Output Format (JSON Only)\\
\{\\
  "error\_type": "Selected error type category",\\
  "diagnosis": "Evaluation about the Step to evaluate.",\\
  "guidance": "Instruction for immediate single next step."\\
\}\\

\# Few-shot Demonstrations

[EXAMPLES...]
\end{tcolorbox}
\caption{Evaluation prompt.}
\label{fig:evaluation_prompt}
\end{figure*}

\begin{figure*}[t]
\begin{tcolorbox}[
    colback=gray!5,
    colframe=black,
    title=Final Answer Generation Prompt,
    fonttitle=\bfseries,
    fontupper=\small,
    sharp corners,
    boxrule=0.5pt
]
You are an expert answering agent.\\
The reasoning process is complete. Your task is to formulate the FINAL ANSWER based on the provided history.\\

INSTRUCTIONS:\\
1. Do not generate any new reasoning steps.\\
2. Directly output the final answer.\\
3. YOU MUST USE THE FOLLOWING FORMAT:
\#\#\#\#ANSWER: your\_final\_answer\_here (Final Answer)
\end{tcolorbox}
\caption{Final Answer Generation prompt.}
\label{fig:final_answer_generation_prompt}
\end{figure*}

\begin{figure*}[t]
\begin{tcolorbox}[
    colback=gray!5,
    colframe=black,
    title=LLM Judge Prompt,
    fonttitle=\bfseries,
    fontupper=\small,
    sharp corners,
    boxrule=0.5pt
]
You are an expert evaluator for a Question Answering task.\\
Your goal is to determine if the 'Generated Answer' is correct based on the 'Ground Truth List'.\\

**Evaluation Criteria:**\\
1. **Semantic Equivalence:** If the 'Generated Answer' refers to the same real-world entity, concept, or event as **ANY** item in the 'Ground Truth List', it is "correct". This includes aliases, abbreviations, and common synonyms.\\
2. **Granularity \& Hierarchy:** Accept answers that are factually accurate but differ in specificity or granularity, provided they refer to the same location or entity.\\
    - **Geographic Inclusion:** Accept constituent countries, states, or specific locations if they are part of the broader Ground Truth entity (e.g., "England" is correct for "United Kingdom"; "New York" is correct for "USA" if the context implies origin).\\
    - **Specificity:** Accept broader correct terms if they encompass the specific Ground Truth (e.g., "UK" is correct for "England" if the question asks for country).\\
3. **Logical Entailment \& Event Description:** Accept answers that describe the same event or fact using different but factually compatible attributes.\\
    - **Cause vs. Nature:** If Ground Truth specifies the mechanism (e.g., "shot") and Generated Answer specifies the nature of the event (e.g., "homicide" or "murder"), and both describe the same factual occurrence, it is "correct".\\
    - **Implication:** If the Generated Answer logically implies the Ground Truth or vice versa in the given context (e.g., "shot by father" implies "killed by family member"), it is "correct".\\
4. **Robustness:** Ignore minor formatting, casing, punctuation, or conversational fillers (e.g., "The answer is...").\\
5. **Contextual Correctness:** If the answer is factually different, references a completely distinct entity, or introduces contradictory information compared to the ground truth (e.g., "died of old age" vs "shot"), it is "wrong".\\

**Output Format:**\\
You must output ONLY a valid JSON object with exactly two keys: "is\_correct" and "reasoning". Do not include any markdown styling or extra text.\\
- "is\_correct": Must be either "correct" or "wrong".\\
- "reasoning": A brief, high-density explanation of why the answer was judged this way, specifically mentioning any logical entailment or hierarchy logic if applicable.\\

[EXAMPLES...]

\end{tcolorbox}
\caption{LLM judge prompt.}
\label{fig:llm_judge_prompt}
\end{figure*}

\begin{figure*}[t]
\begin{tcolorbox}[
    colback=gray!5,
    colframe=black,
    title=Plan Generation Prompt (2Wiki),
    fonttitle=\bfseries,
    fontupper=\small,
    sharp corners,
    boxrule=0.5pt
]
You are an expert AI assistant specializing in query analysis and multi-hop reasoning. 
Your task is to analyze a given question and generate the minimal, step-by-step reasoning plan required to answer it.\\

**Instructions:**\\
1. Read the user's question and identify the two entities being compared (Entity A and Entity B) if it is a comparison question.\\
2. Identify the specific attribute being compared (e.g., country, release date, lifespan).\\
3. Determine the correct reasoning plan based on the question type:\\
   * **If the question is a Comparison:**\\
     * **Plan 1 (Simple Attribute Comparison):** The attribute is a single value that can be directly looked up (e.g., country, date, nationality). This plan involves 3 steps: Find A, Find B, Compare.\\
     * **Plan 2 (Calculated Attribute Comparison):** The attribute must be calculated from other facts (e.g., lifespan). This plan involves 5 steps: Find facts for A, Calculate attribute for A, Find facts for B, Calculate attribute for B, Compare.\\
     * **Plan 3 (Set Attribute Comparison):** The attribute is a list whose size must be counted (e.g., number of professions, number of directors). This plan involves 5 steps: Find list for A, Count list for A, Find list for B, Count list for B, Compare.\\
     * **Plan 4 (Bridge Comparison):** The comparison is about an attribute of a "bridge entity" (e.g., director) connected to the main entities (e.g., films). This plan involves 5 steps: Find bridge entity A, Find attribute of bridge entity A, Find bridge entity B, Find attribute of bridge entity B, Compare the attributes to determine which main entity satisfies the condition.\\
   * **If the question is a Multi-hop Fact Retrieval:**\\
     * **Plan 5 (Sequential Fact Retrieval):** Generate the sequential chain of attribution and logical steps needed to find the answer.\\
4. Generate the reasoning plan as a numbered list.\\
5. Label each step with its type: `(Attribution)` for finding information or `(Logical)` for reasoning, calculating, or comparing.\\
6. **CRITICAL (Dependencies):** If a step (e.g., Step 2) uses the information found in a previous step (e.g., Step 1), you **must** explicitly refer to it. (e.g., "Find the father of the person from Step 1.")\\
7. **CRITICAL (Atomicity):** Do NOT combine multiple actions into a single step. Specifically, do not find an entity and its attribute simultaneously (e.g., "Find the director and their birth date"). This must be split into two steps: first identify the entity, then find the attribute.\\
8. The plan should only contain the intermediate reasoning steps, not the final answer.\\
9. **CRITICAL (Final Step):** All reasoning plans must end with a `(Logical)` step, which compares, calculates, or identifies the final piece of information required by the question. This step must explicitly determine the specific entity or value requested by the question (e.g., if the question asks 'Which film...', the final step must be 'determine which film ...').\\
10. **CRITICAL (Format):** You MUST output **only** the list in the specified format: `[Step 1: ..., Step 2: ..., ...]`. Do not include *any* other text, JSON formatting, explanations, or conversational chat before or after the list.

---

**Examples:**

[EXAMPLES...]
\end{tcolorbox}
\caption{Plan generation prompt for 2WikiMultiHop dataset.}
\label{fig:plan_generation_prompt_2wiki}
\end{figure*}

\begin{figure*}[t]
\begin{tcolorbox}[
    colback=gray!5,
    colframe=black,
    title=Plan Generation Prompt (HotpotQA),
    fonttitle=\bfseries,
    fontupper=\small,
    sharp corners,
    boxrule=0.5pt
]
You are an expert AI assistant specializing in query analysis and multi-hop reasoning. 
Your task is to analyze a given question and generate the minimal, step-by-step reasoning plan required to answer it.\\

**Instructions:**\\
1. Read the user's question and determine its primary type:\\
   * **Fact Retrieval (Sequential/Compositional/Inference):** The question asks for a single fact (Who/What/When/Where) which requires finding intermediate facts first (e.g., "Who is the mother of the director of..."). This is **Plan 4**.\\
   * **Comparison:** The question compares two or more entities (e.g., "Who was born first, A or B?", "Which has more..."). This is **Plan 1, 2, or 3**.\\
2. **If the question is a Comparison:**\\
   * **Plan 1 (Simple Attribute Comparison):** The attribute is a single value that can be directly looked up (e.g., country, date, nationality). This plan involves 3 steps: Find A, Find B, Compare.\\
   * **Plan 2 (Calculated Attribute Comparison):** The attribute must be calculated from other facts (e.g., lifespan, duration). This plan involves 5 steps: Find facts for A, Calculate attribute for A, Find facts for B, Calculate attribute for B, Compare.\\
   * **Plan 3 (Set Attribute Comparison):** The attribute is a list whose size must be counted (e.g., number of professions, number of members). This plan involves 5 steps: Find list for A, Count list for A, Find list for B, Count list for B, Compare.\\
3. **If the question is a Fact Retrieval:**
   * **Plan 4 (Sequential Fact Retrieval):** Generate the sequential chain of attribution and logical steps needed to find the answer.\\
4. **CRITICAL (Minimal Atomic Plan):** The plan must be **minimal** and **atomic**.\\ 
   * A search for an entity with multiple constraints (e.g., "the 2011 film scored by Chris Bacon that is based on Shakespeare") must be a **single (Attribution) step**, not multiple filtering steps.\\
   * Descriptive clauses (e.g., "Ku Hye-sun, who appeared in...") are constraints for the *first* search step. **DO NOT** create separate steps for "confirmation" or "verification" of this descriptive information.\\
5. Generate the reasoning plan as a numbered list.\\
6. Label each step with its type: `(Attribution)` for finding information or `(Logical)` for reasoning, calculating, or comparing.\\
7. **CRITICAL (Dependencies):** If a step (e.g., Step 2) uses the information found in a previous step (e.g., Step 1), you **must** explicitly refer to it. (e.g., "Find the director of the film found in Step 1.")\\
8. **CRITICAL (Final Step):** All reasoning plans must end with a `(Logical)` step, which compares, calculates, or identifies the final piece of information required by the question. This step must explicitly determine the specific entity or value requested by the question (e.g., if the question asks 'Which film...', the final step must be 'determine which film ...').\\
9. Output the steps in a **list format**: `[Step 1: ..., Step 2: ..., ...]`\\
10. **CRITICAL (Format):** You MUST output **only** the list in the specified format. Do not include *any* other text, JSON formatting, explanations, or conversational chat before or after the list.

---

**Examples:**

[EXAMPLES...]
\end{tcolorbox}
\caption{Plan generation prompt for HotpotQA dataset.}
\label{fig:plan_generation_prompt_hotpotqa}
\end{figure*}

\begin{figure*}[t]
\begin{tcolorbox}[
    colback=gray!5,
    colframe=black,
    title=Plan Generation Prompt (MuSiQue),
    fonttitle=\bfseries,
    fontupper=\small,
    sharp corners,
    boxrule=0.5pt
]
You are an expert AI assistant specializing in query analysis and reasoning plan generation.
Your task is to translate a given Question Decomposition (a list of sub-questions/steps) into a formal, step-by-step reasoning plan.\\

The user will provide two inputs:\\
**Question:** The original multi-hop question.\\
**Question Decomposition:** A numbered list of sub-questions labeled as Q1, Q2, Q3, etc.
Each sub-question describes one reasoning step, where later steps (e.g., Q2, Q3) may depend on previous results using references like "\#1" (meaning the answer from Q1) or "\#2" (meaning the answer from Q2).\\

Your task is to convert this Question Decomposition into a reasoning plan with (Attribution) and (Logical) steps.\\

**Instructions:**\\

1. Read the Question Decomposition list carefully. Each line will be labeled as Q1, Q2, Q3, etc. Treat each "Qn" as one reasoning step in sequence.\\
2. For each item in the Question Decomposition list, create one corresponding (Attribution) step.\\
Example: A decomposition step like 'Q1: Keep the Faith >> performer' must be translated into Step 1: Find the performer of "Keep the Faith". (Attribution).\\
3. **CRITICAL (Dependencies):** Translate the dependencies exactly.
A decomposition step like 'Q2: \#1 >> record label' must be translated to Step 2: Find the record label for the entity found in Step 1. (Attribution).
A decomposition step like 'Q3: \#2 >> genre' must be translated to Step 3: Find the genre of the entity found in Step 2. (Attribution).\\
4. **CRITICAL (Final Step):** After translating all steps from the Question Decomposition, add one final (Logical) step.
This final (Logical) step must identify the result of the last attribution step as the answer, based on what the original Question was asking.
Example: If the last attribution step was Step 3: Find the genre... and the original Question was "What is the... genre...", the final step must be "Step 4: Identify the genre found in Step 3 as the answer. (Logical)".\\
5. **CRITICAL (Format):** Output the plan as a numbered list in a single list format: [Step 1: ..., Step 2: ..., ...]\\
6. **CRITICAL (Output Only):** You MUST output only the list in the specified format. Do not include any other text, JSON formatting, explanations, or conversational chat before or after the list.

---

**Examples:**

[EXAMPLES...]
\end{tcolorbox}
\caption{Plan generation prompt for MuSiQue dataset.}
\label{fig:plan_generation_prompt_musique}
\end{figure*}

\begin{figure*}[t]
\begin{tcolorbox}[
    colback=gray!5,
    colframe=black,
    title=Ideal Reasoning Generation Prompt,
    fonttitle=\bfseries,
    fontupper=\small,
    sharp corners,
    boxrule=0.5pt
]
You are an expert AI assistant specializing in multi-hop reasoning. 
Your task is to generate the ideal, step-by-step reasoning that correctly follows a given reasoning plan, using only the provided ground truth context.\\

**Instructions:**\\
1. You will be given a `Question`, a `Ground Truth Context`, and a `Reasoning Plan` (list of instructions).\\
2. Your goal is to "execute" the `Reasoning Plan` to generate the final reasoning steps.\\
3. You must generate exactly one reasoning step for each instruction in the `Reasoning Plan`.\\
4. **CRITICAL (Context):** Your reasoning *must* be based *only* on the facts provided in the `Ground Truth Context`. Do not use any external knowledge.\\
5. **CRITICAL (Citation):** For all `(Attribution)` steps, you **must** explicitly cite the passage number in the text ("According to Passage 1, ...").\\
6. **CRITICAL (Tags):** Each reasoning step you generate *must* end with the exact `(Attribution)` or `(Logical)` tag that appears in the corresponding plan step.\\
7. **CRITICAL (Dependencies):** When a plan step has a dependency (e.g., "...from Step 1"), your reasoning step must clearly show this (e.g., "The father of Bengt Snivil (from Step 1) is...").\\
8. **CRITICAL (Index Mapping):**
   - For **(Attribution)** steps: `supporting\_index` must be the integer index of the Passage used (e.g., 1 for Passage 1).
   - For **(Logical)** steps: `supporting\_index` must be a list of indices of the previous steps used as evidence (e.g., [2, 4]).\\
9. **CRITICAL (Format):** You MUST output **only** a valid JSON-style list of dictionaries. Do not include markdown code blocks (```json), backticks, or any conversational text. Start directly with '[' and end with ']'.

---

**Examples:**

[EXAMPLES...]
\end{tcolorbox}
\caption{Ideal reasoning generation prompt.}
\label{fig:ideal_reasoning_generation_prompt}
\end{figure*}

\begin{figure*}[t]
\begin{tcolorbox}[
    colback=gray!5,
    colframe=black,
    title=Error Injection Prompt,
    fonttitle=\bfseries,
    fontupper=\small,
    sharp corners,
    boxrule=0.5pt
]
You are an expert in logical reasoning, tasked with intentionally introducing a specific logical error into a reasoning steps.\\

Your Goal: Replace a single, correct reasoning step with an \{error\_type\} error.\\

Error Definition:

[ERROR\_DEFINITION]\\

Input Format:\\
You will receive:\\
1. Question: The user's original question.\\
2. Retrieved Passages: Contextual information.\\
3. Ideal Reasoning Steps: The correct, multi-step reasoning.\\
4. Target Step to Corrupt: The specific step from the ideal steps that you must replace. (ex. Step 2)\\

Output Format:\\
- You MUST output only the single, new, erroneous reasoning step.\\
- The new step MUST be formatted exactly like the target step, including the "Step X:" prefix and the "(Label)" suffix.\\

---\\
EXAMPLES\\
---\\

[Examples...]
\end{tcolorbox}
\caption{Error Injection prompt.}
\label{fig:error_injection_prompt}
\end{figure*}

\begin{figure*}[t]
\begin{tcolorbox}[
    colback=gray!5,
    colframe=black,
    title=Triple Extraction Prompt,
    fonttitle=\bfseries,
    fontupper=\small,
    sharp corners,
    boxrule=0.5pt
]
You are an expert Knowledge Graph Engineer and Computational Linguist. Your task is to extract all factual triples from the provided text passage to construct a comprehensive Knowledge Graph.\\

\#\#\# CRITICAL INSTRUCTION: COREFERENCE RESOLUTION (DECONTEXTUALIZATION)\\
Before extracting any triples, you must strictly perform **Coreference Resolution** internally.\\
- **Never use pronouns** (e.g., "he", "she", "it", "they", "this", "that", "the film", "the company") as the Subject or Object in the output triples.\\
- You must identify the specific entity the pronoun refers to from the context and replace the pronoun with the **full canonical name** of that entity.\\
- Example: If the text says "Barack Obama: Barack Obama was born in Hawaii. He served as the 44th president.", the extracted triple must be `["Barack Obama", "served\_as", "44th president"]`, NOT `["He", "served\_as", "44th president"]`.\\

\#\#\# EXTRACTION GUIDELINES\\
1. **Atomic Triples:** Break down complex sentences into atomic facts (Subject, Predicate, Object).\\
2. **Predicate Standardization:** Use clear, concise, and meaningful predicates (e.g., "born\_in", "directed\_by", "spouse", "occupation"). Avoid vague verbs like "is" or "has" if a more specific relation exists.\\
3. **Attribute Extraction:** Extract numerical values, dates, and specific roles as Objects.\\
   - Dates should be in `YYYY-MM-DD` format if possible.\\
   - Numbers should be kept as raw values suitable for comparison.\\

\#\#\# OUTPUT FORMAT\\
Output **ONLY** a standard JSON list of lists. Do not include any explanation or markdown formatting outside the JSON.
Each inner list must strictly follow the order: `[Subject, Predicate, Object]`. The Subject is the entity being described.\\

JSON Format:
  
[["Entity Name", "Relation/Attribute", "Entity Name/Value"],
  ["Entity Name", "Relation/Attribute", "Entity Name/Value"],
  ...]\\

\#\#\# EXAMPLES

[EXAMPLES...]
\end{tcolorbox}
\caption{Triple extraction prompt.}
\label{fig:triple_extraction_prompt}
\end{figure*}

\begin{figure*}[t]
\begin{tcolorbox}[
    colback=gray!5,
    colframe=black,
    title=Triple Gleaning Prompt,
    fonttitle=\bfseries,
    fontupper=\small,
    sharp corners,
    boxrule=0.5pt
]
You are a meticulous Knowledge Graph Auditor and QA Specialist.
Your goal is to review the extraction results from a previous step and identify **MISSING** factual triples that were overlooked.\\

\#\#\# INPUTS\\
You will be provided with:\\
1. **Original Text Passage**: The source text.\\
2. **Existing Triples**: A list of triples already extracted from the text.\\

\#\#\# TASK\\
Compare the **Original Text** against the **Existing Triples**. Extract **NEW** factual triples that are present in the text but **NOT** in the Existing Triples list.\\

\#\#\# CRITICAL RULES\\
1. **NO DUPLICATES**: Do not output any triple that is semantically identical to one in the "Existing Triples" list.\\
2. **Focus on Detail**: Look specifically for:
   - Secondary entities mentioned in the text (not just the main subject).\\
   - Specific dates, numbers, or measurements previously missed.\\
   - Relationships between secondary entities.\\
   - Adjectives or roles acting as attributes (e.g., "Polish-French" -> nationality).\\
3. **Coreference Resolution (Mandatory)**: Just like the first step, you MUST resolve pronouns (he, she, it, they) to their full canonical entity names. **Never** use pronouns in your output.\\
4. **Atomic \& Standardized**: Follow the same extraction standards (Atomic facts, standardized predicates).\\
5. **Aliases \& Alternate Names:**\\
   - If an entity has multiple names (e.g., "better known as", "born as", "pseudonym", "nickname"), you MUST extract a triple linking them.\\
   - Use predicates like `same\_as`, `alias`, `birth\_name`, or `alternative\_name`.\\
   - Example: "Donna Paige Helmintoller, better known as Paige O'Hara" -> Extract `["Donna Paige Helmintoller", "same\_as", "Paige O'Hara"]`.\\

\#\#\# OUTPUT FORMAT\\
Output **ONLY** a standard JSON list of lists containing the **NEWLY EXTRACTED** triples.
If no new triples are found, output an empty list `[]`.\\

JSON Format:

[["Subject", "Predicate", "Object"],
  ["Subject", "Predicate", "Object"],
  ...]
\end{tcolorbox}
\caption{Triple gleaning prompt.}
\label{fig:triple_gleaning_prompt}
\end{figure*}

\begin{figure*}[t]
\begin{tcolorbox}[
    colback=gray!5,
    colframe=black,
    title=Entity Resolution Prompt,
    fonttitle=\bfseries,
    fontupper=\small,
    sharp corners,
    boxrule=0.5pt
]
You are an expert Knowledge Graph Engineer specializing in Entity Resolution (Coreference Resolution).
Your task is to identify and group entities that refer to the **same real-world object** (Named Entities) from the provided list of entities and context triples.\\

\#\#\# INSTRUCTIONS\\
1. **Analyze Entities:** Review the provided list of entities.\\
2. **Check Context:** Use the "Context Triples" to verify if similar names refer to the exact same unique entity.\\
3. **Group Synonyms:** Group variations (abbreviations, acronyms, partial names, typos) of the same **Specific Named Entity** into a single list.\\
4. **Filter Noise:** **STRICTLY EXCLUDE** generic terms, common nouns, numbers, dates, and abstract concepts.\\

\#\#\# OUTPUT FORMAT\\
Output **ONLY** a JSON list of lists. Each inner list represents a group of synonymous entities.\\
Example: `[["USA", "United States", "U.S."], ["JFK", "John F. Kennedy"], ["The Green Album", "Green"]]`\\

\#\#\# RULES\\
1. **Target Proper Nouns Only:** Only group Specific People, Places, Organizations, Events, and Works.\\
2. **Exclude Generics:** Do NOT include or group common nouns or roles.\\
   - **Bad:** "president", "mother", "clothes", "the album", "the band", "different stage names", "two sons".\\
   - **Good:** "President Obama", "Mother Teresa", "The White Album", "System 7".\\
3. **Exclude Literals:** Do NOT include or group Values.\\
   - **Dates:** "2020-01-01", "1990s", "May 5th".\\
   - **Numbers:** "3", "100", "first", "one".\\
   - **Measurements:** "3 years", "10kg", "50\%".\\
4. **Avoid Super Nodes:** Never group a specific entity with a generic category (e.g., DO NOT group `["Barack Obama", "President"]`). "President" is a title/role, not the person's unique identifier.\\
5. **Context is King:** If "Jordan" appears in a triple about basketball and another about the Middle East, DO NOT group them.
\end{tcolorbox}
\caption{Entity resolution prompt.}
\label{fig:entity_resolution_prompt}
\end{figure*}

\begin{figure*}[t]
\begin{tcolorbox}[
    colback=gray!5,
    colframe=black,
    title=Logical Path Discovery Prompt,
    fonttitle=\bfseries,
    fontupper=\small,
    sharp corners,
    boxrule=0.5pt
]
You are an expert Knowledge Graph Reasoner.
Your task is to identify the **Reasoning Evidence (Subgraph)** required to answer a Multi-hop Question using the provided Normalized Triples.\\

\#\#\# GOAL\\
Select the **minimal set of triples** that acts as the necessary premises to logically derive the **Ground Truth Answer**.\\

\#\#\# UNDERSTANDING REASONING PATTERNS\\
The "Reasoning Path" is NOT always a single connected chain. It can be:\\
1.  **Sequential (Bridge):** (e1, r1, e2) -> (e2, r2, e3)\\
    * *Ex:* "When is the birthplace of the director of the movie Parasite?"\\
2.  **Parallel (Comparison/Intersection):** (e1, r1, v1) and (e2, r1, v2)\\
    * *Ex:* "Are Obama and Trump born in the same country?" (Requires birthplaces of BOTH persons, even if they are disjoint nodes).\\
    * *Ex:* "Who is older, A and B?" (Requires birthdates of BOTH persons).\\

\#\#\# INSTRUCTIONS\\
1.  **Analyze the Logic:** Determine if the question requires a chain of facts, a comparison of two facts, or an intersection of attributes.\\
2.  **Select Evidence:** Pick ONLY the triples necessary to support the reasoning.\\
    * **Ignore Direction:** Treat relations as bidirectional if semantically appropriate (e.g., `["Movie", "directed\_by", "Director"]` supports finding the director).\\
    * **Semantic Match:** Match predicates loosely (e.g., "born\_in" is equal to "place of birth").\\
3.  **Verify \& Explain:**\\
    * Do these selected triples logically lead to the Ground Truth Answer?\\
    * If the required triples are missing (e.g., one side of a comparison is missing), mark as invalid.\\

\#\#\# OUTPUT FORMAT\\
Output **ONLY** a JSON object:\\
\{\\
  "is\_valid": true, // true if ALL necessary evidence is present to derive the Ground Truth Answer\\
  "reasoning\_path": [ // List of selected triples. Can be disjoint (e.g., two separate birth dates).
    ["Subject1", "Predicate1", "Object1"],
    ["Subject2", "Predicate2", "Object2"],
    ...
  ],\\
  "explanation": "Briefly explain the logic (e.g., 'Found birth dates for both persons to compare them')."\\
\}\\

\#\#\# EXAMPLES\\

[Examples...]
\end{tcolorbox}
\caption{Logical path discovery prompt.}
\label{fig:logical_path_discovery_prompt}
\end{figure*}



\end{document}